\documentclass[letterpaper]{article} 
\usepackage{aaai25}  
\usepackage{times}  
\usepackage{helvet}  
\usepackage{courier}  
\usepackage[hyphens]{url}  
\usepackage{graphicx} 
\usepackage{multirow}
\usepackage{booktabs}
\usepackage{dashrule}
\PassOptionsToPackage{table,dvipsnames}{xcolor}
\usepackage[table,xcdraw]{xcolor}
\usepackage{natbib}  
\usepackage{caption} 
\usepackage{lipsum}
\usepackage{lineno}
\usepackage{soul}
\usepackage[most]{tcolorbox}
\newcommand{\tboxcomment}[1]{\textcolor{gray}{\# #1}}
\newcommand{\highlight}[2]{\sethlcolor{#1}\hl{#2}}
\definecolor{cyan10}{HTML}{E5F6FF}
\definecolor{cyan20}{HTML}{BAE6FF}
\definecolor{cyan60}{HTML}{0072c3}
\definecolor{cyan70}{HTML}{00539a}
\definecolor{teal10}{HTML}{D9FBFB}
\definecolor{teal20}{HTML}{9EF0F0}
\definecolor{teal60}{HTML}{007d79}
\definecolor{orange10}{HTML}{FFF2E8}
\definecolor{orange20}{HTML}{FFD9BE}
\definecolor{orange60}{HTML}{ba4e00}
\definecolor{blue10}{HTML}{EDF5FF}
\definecolor{blue20}{HTML}{D0E2FF}
\definecolor{magenta10}{HTML}{FFF0F7}
\definecolor{magenta20}{HTML}{FFD6E8}
\definecolor{magenta30}{HTML}{ffafd2}
\definecolor{magenta50}{HTML}{ee5396}
\definecolor{magenta60}{HTML}{d02670}
\definecolor{magenta70}{HTML}{9f1853}
\definecolor{purple10}{HTML}{F6F2FF}
\definecolor{purple20}{HTML}{E8DAFF}
\definecolor{purple30}{HTML}{d4bbff}
\definecolor{purple70}{HTML}{8a3ffc}
\definecolor{rose10}{HTML}{FCF2ED}
\definecolor{rose20}{HTML}{F9D9D1}
\definecolor{red10}{HTML}{FFF1F1}
\definecolor{red20}{HTML}{FFD7D9}
\definecolor{green10}{HTML}{DEFBE6}
\definecolor{green20}{HTML}{A7F0BA}
\definecolor{yellow10}{HTML}{fcf4d6}
\definecolor{yellow20}{HTML}{fddc69}
\definecolor{gray20}{HTML}{e0e0e0}
\definecolor{gray30}{HTML}{c6c6c6}
\definecolor{gray40}{HTML}{a8a8a8}
\definecolor{gray80}{HTML}{393939}

\newcommand{\badval}{\cellcolor[HTML]{FFCCC9}}
\newcommand{\goodval}{\cellcolor[HTML]{9AFF99}}

\usepackage{todonotes}

\usepackage{amsthm}
\usepackage{amsmath}
\usepackage{amsfonts}
\usepackage{amssymb}

\theoremstyle{plain}
\newtheorem{theorem}{Theorem}[section]

\theoremstyle{definition}
\newtheorem{definition}[theorem]{Definition}

\theoremstyle{remark}

\usepackage{algpseudocodex}
\usepackage{algorithm}
\usepackage{booktabs}
\usepackage{newfloat}
\usepackage{listings}
\usepackage[dvipsnames]{xcolor}
\definecolor{codegreen}{rgb}{0,0.6,0}
\definecolor{codegray}{rgb}{0.5,0.5,0.5}
\definecolor{codepurple}{rgb}{0.58,0,0.82}
\definecolor{backcolour}{rgb}{0.95,0.95,0.92}

\DeclareCaptionStyle{ruled}{labelfont=normalfont,labelsep=colon,strut=off} 
\floatstyle{ruled}
\newfloat{listing}{tb}{lst}{}
\floatname{listing}{Listing}
\usepackage{hyperref} 
\nocopyright 

\title{Chasing Progress, Not Perfection: Revisiting Strategies for\\End-to-End LLM Plan Generation}
\author {
    Sukai Huang,
    Nir Lipovetzky and 
    Trevor Cohn\thanks{Now at Google DeepMind}
}
\affiliations {
    School of Computing and Information Systems, The University of Melbourne, Australia\\
    sukaih@student.unimelb.edu.au, \{nir.lipovetzky,
    trevor.cohn\}@unimelb.edu.au \\
}

\begin{document}


\maketitle 

\begin{abstract}
    The capability of Large Language Models (LLMs) to plan remains a topic of debate. Some critics argue that strategies to boost LLMs' reasoning skills are ineffective in planning tasks, while others report strong outcomes merely from training models on a planning corpus. This study reassesses recent strategies by developing an end-to-end LLM planner and employing diverse metrics for a thorough evaluation. We find that merely fine-tuning LLMs on a corpus of planning instances does not lead to robust planning skills, as indicated by poor performance on \emph{out-of-distribution} test sets. At the same time, we find that various strategies, including \emph{chain-of-thought}, do enhance the probability of a plan being \emph{executable}. This indicates progress towards better plan quality, despite not directly enhancing the final \emph{validity rate}. Among the strategies we evaluated, reinforcement learning with our novel `Longest Contiguous Common Subsequence' reward emerged as the most effective, contributing to both plan executability and validity. Overall, our research addresses key misconceptions in the LLM-planning literature; we validate incremental progress in plan executability, although plan validity remains a challenge. Hence, future strategies should focus on both these aspects, drawing insights from our findings.
\end{abstract}

\section{Introduction}
\label{sec:intro}

While latest Large Language Models (LLMs) have shown promise in areas like grade school math and code generation \citep{liu2023tinygsm, Ye2024PhysicsOL, kumar2024training, madaan2024self}, their effectiveness in solving planning tasks remains a contentious issue. Numerous studies have voiced skepticism, questioning whether LLMs can truly conduct deliberative reasoning beyond statistical pattern matching \citep{Huang2023ASO, kambhampatiposition2024, valmeekam2024planning}. However, the landscape is not uniformly skeptical, some studies have also made provocative claims that LLMs, when fine-tuned on paired datasets of planning problem descriptions and their corresponding plans, can generate correct plan sequences for new problems within the same domains \citep{shah2024causal, rossetti2024learning}.

Existing literature on \emph{LLMs planning} shows a clear dichotomy, likely due to limitations in evaluation methodologies on both sides of the debate, as shown below:

\noindent\textbf{1. Lack of OOD Evaluation:} Optimistic studies often overlook out-of-distribution (OOD) evaluation, leading to models potentially simply recalling and adapting partial plan traces from the training data, rather than demonstrating genuine sequential reasoning \citep{mirzadeh2024gsm}. For instance, \citet{shah2024causal} evaluated their model on Sudoku puzzles of the same size as those in the training set, potentially allowing the model to reuse familiar grid patterns. Similarly, \citet{rossetti2024learning} evaluated their fine-tuned LLM planner on test instances generated using the same PDDL generator configurations as the training data. This lack of OOD evaluation can lead to an overestimation of the model's planning capabilities. 

\begin{figure*}[t]
    \centering
    \includegraphics[width=\textwidth]{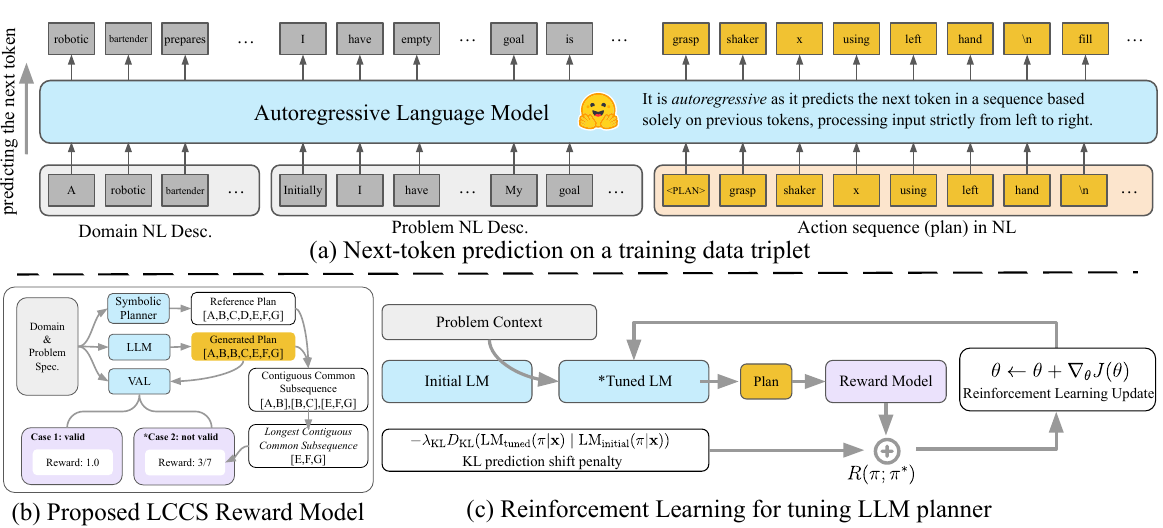}
    \caption{\emph{(a) Next-token prediction:} The LLM is trained to predict the next token in corpus containing domain and problem details and their plans, proceeding left-to-right. (b) \emph{Proposed LCCS Reward:} We use the length of LCCS as an auxiliary reward signal for RL. It provides granular feedback to fill the sparse reward gap. (See Section~\ref{subsec:rl}). (c) \emph{Reinforcement Learning:}  RL with LCCS Reward is shown to be the most effective strategy for enhancing end-to-end LLM planners among all tested strategies.}
    \label{fig:next_token_prediction_illu}
\end{figure*}

\noindent\textbf{2. Insufficient Analysis of Failure Modes:} Pessimistic studies (e.g., \citet{valmeekam2024planning}) typically focus solely on plan \emph{validity}, an extremely stringent criterion that requires the entire plan to be perfectly correct. This restrictive metric fails to capture incremental improvements in planning capabilities. Moreover, these studies rarely conduct diagnostic experiments to identify specific reasoning bottlenecks, leading to a puzzling enigma in the planning community -- Why do strategies that enhance reasoning capabilities in other tasks (e.g., math-solving \citep{wei2022chain}) appear ineffective in planning tasks? For instance, \citet{stechly2024chain} found that chain-of-thought prompting did not improve the validity rate of LLM planners. By not considering more granular metrics and failing to investigate where the strategy fails, researchers miss opportunities to understand how different strategies contribute and which aspects of reasoning need the most attention.

To address these gaps, we \textbf{reassess} various strategies for training LLMs in \emph{end-to-end plan generation}, where the LLM itself functions as a \textbf{black-box planner} with \textbf{no} explicitly modeled \emph{search process}. These strategies include \emph{permutation} \citep{allen2023physics}, \emph{chain-of-thought} \citep{wei2022chain, yao2024tree}, \emph{self-correction} \citep{madaan2024self, Ye2024PhysicsOL, kumar2024training}, and \emph{reinforcement learning} (RL) \citep{liu2024deepseek}. Our evaluation extends beyond the plan \emph{validity} metric to include plan \emph{executability\footnote{An executable plan is one where the preconditions of all actions are satisfied by the state at execution (See Appendix~\ref{app:terminology}).}} and employs a comprehensive suite of diagnostic experiments to identify where each strategy succeeds or fails. Moreover, OOD test sets are deliberately designed to evaluate the model's generalization capabilities in planning. 

We hope our work will address misconceptions in the LLM planning literature and provide a better understanding of the inherent planning capabilities of LLMs' \emph{next-token prediction} mechanism. Our contributions are as follows: (1) We challenge the claim that \emph{fine-tuning} LLMs simply on datasets containing problem contexts and reference plans acquire robust planning skills, by demonstrating their failure on OOD test sets. (2) We show that strategies like \emph{CoT} lead to incremental improvements in plan quality by enhancing plan \emph{executability}, even if they do not directly increase \emph{validity} rates. (3) We show that RL with our proposed `\textsc{LCCS}' reward emerges as the most effective strategy. In particular, it improves plan \emph{validity} by 7\% and \emph{executability} by 9\% in longer planning problems.

\section{Background \& Related Work}

\textbf{Classical Planning.}\quad We assume that readers are familiar with the standard planning language PDDL for representing deterministic, fully observable planning problems. A classical planning problem is a pair $P = \langle D, I \rangle$, where $D$ represents the planning domain, consisting of a set of predicate symbols and action schemas; $I$ denotes the problem instance, which includes the objects, initial state and goal. A sequence of grounded actions $\pi = (a_0, a_1, \ldots, a_n)$ is a \emph{valid} plan \emph{iff} it is \emph{executable} and the goal $G$ holds in the final state after executing the plan.

\noindent \textbf{Next-token prediction.}\quad We fine-tune an end-to-end LLM-planner, $\text{LM}_\theta$, using \emph{next-token prediction} on a text sequence $\mathbf{x} = \langle \mathcal Z(D), \mathcal Z(I), \mathcal Z(\pi) \rangle$, where $Z(\cdot)$ refers to the serialization of the PDDL elements into natural language (as illustrated in Figure~\ref{fig:next_token_prediction_illu} part a). Let $\mathbf{x}_{<i}$ denote the first $i-1$ tokens of sequence $\mathbf{x}$, and $x_i$ be the $i$-th token. Then, $\text{LM}_\theta(\hat x_i =x_i \mid \mathbf{x}_{<i})$ indicates the probability the model predicts the $i$-th token $\hat x_i$, to be equal to the actual token $x_i$, conditioned on the preceding tokens $\mathbf{x}_{<i}$. This is also known as \emph{autoregressive modeling}. The training objective is to maximize the likelihood of the joint distribution of the training corpus $\mathcal X$, expressed as follows: $J(\theta) = \max_\theta \mathbb{E}_{\mathbf{x} \sim \mathcal X} \left[ \sum_{i=1}^{|\mathbf{x}|} \log \text{LM}_\theta(\hat x_i =x_i \mid \mathbf{x}_{<i}) \right]$. Note that this approach trains the model to predict not only tokens in the output (i.e., the generated plans) but also tokens in the input query (i.e., the domain and problem description). 

\noindent\textbf{Scope.}\quad We examine the planning skills of LLMs in the paradigm of \textbf{end-to-end plan generation}. Our focus excludes certain approaches, such as the LLM-Modulo framework, which relies on an external verifier to validate generated plans \citep{kambhampatiposition2024}, nor the Thought of Search \citep{katz2024thought, cao2024automating}, which bypasses direct long-term planning by prompting LLMs to generate search functions instead of actual plans. We focus solely on paradigms that utilize basic next-token prediction during inference. This excludes hybrid models, such as AlphaMath \citep{chen2024alphamath}, which explicitly model \emph{search process} via Monte Carlo Tree Search (MCTS) to derive solutions. The most closely related work to ours is the \emph{PlanGPT} model \citep{rossetti2024learning}. They demonstrated that fine-tuning LLMs to predict next tokens on plan sequences, without access to the entire search trace, can effectively teach the model to generate plans for new problems. 

However, our work differs in two key aspects: (1) We directly use natural language descriptions instead of PDDL snippets to train the model. This approach aligns more closely with how we expect to interact with LLMs in real-world scenarios. (2) As mentioned in \S~\ref{sec:intro}, we conduct a more comprehensive evaluation to examine the quality of the generated plans and identify the model's reasoning bottlenecks.


\noindent\textbf{Criticism on LLM Planning.}\quad Criticisms of LLMs in planning tasks stem from both theoretical analysis and empirical observations. 
Theoretically, LLMs are machine learning-based \emph{probabilistic models}, and the accuracy of the models' predictions decays exponentially over the length of the sequence, a phenomenon described as ``Snowballing error due to autoregressive modeling'' \citep{bachmann2024pitfalls}. For instance, even with a 99\% correctness rate per step, the probability of a correct sequential prediction drops to about 36.6\% over 100 steps. Therefore, one has to admit that there is no guarantee of soundness in long-term planning tasks due to its probabilistic nature. 

Empirically, numerous studies have demonstrated the struggles of off-the-shelf LLMs in predicting valid plans for long-term tasks \citep{Valmeekam2022PlanBenchAE, valmeekam2023planning, Huang2023ASO}. Even advanced proprietary models like OpenAI o1, which was designed for reasoning tasks, fail on long-term planning \citep{valmeekam2024planning}. In contrast to the pessimistic findings in the literature, we seek to better understand why LLMs often fail in planning tasks and provide insights into how to improve their planning capabilities.

\section{Methodology \& Experimental Design}
\subsection{Extended PlanBench Dataset}
\label{sec:dataset}

\begin{figure}[t]
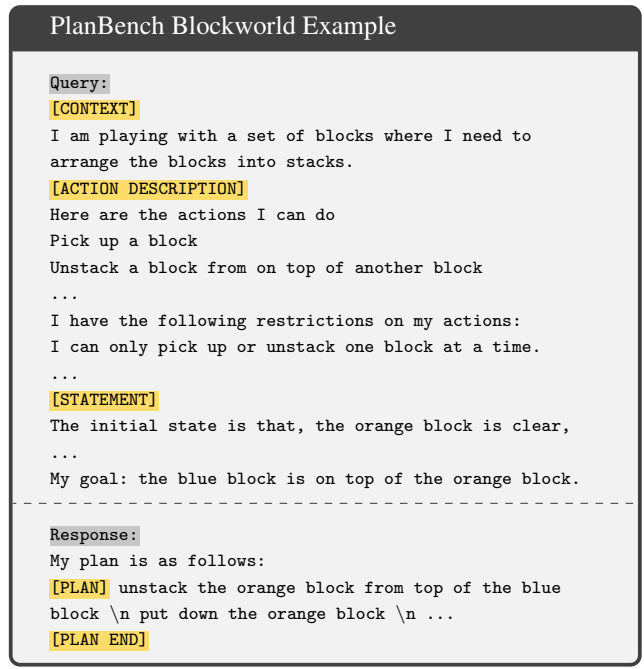

    \centering
    \begin{tcolorbox}[title=PlanBench Blockworld Example, height=8.8cm, fontupper=\scriptsize\ttfamily\fontfamily{cmtt}\selectfont, fontlower=\scriptsize\ttfamily\fontfamily{cmtt}\selectfont]
        \highlight{gray30}{Query:}\\
        \highlight{yellow20}{[CONTEXT]}\\
        I am playing with a set of blocks where I need to arrange the blocks into stacks.\\
        \highlight{yellow20}{[ACTION DESCRIPTION]}\\
        Here are the actions I can do\\
        Pick up a block\\
        Unstack a block from on top of another block\\
        \dots\\
        I have the following restrictions on my actions:\\I can only pick up or unstack one block at a time.\\
        \dots\\
        \highlight{yellow20}{[STATEMENT]}\\
        The initial state is that, the orange block is clear,\\
        \dots\\
        My goal: the blue block is on top of the orange block.
        \vspace{-0.1cm}
        \tcblower
        \highlight{gray30}{Response:}\\
        My plan is as follows:\\
        \highlight{yellow20}{[PLAN]} 
        unstack the orange block from top of the blue block $\backslash$n
        put down the orange block $\backslash$n 
        \dots\\
        \highlight{yellow20}{[PLAN END]}
    \end{tcolorbox}
    \caption{Each planning problem instance in PlanBench is serialized into a single text block that presents the context, action description, initial and goal states, and the plan.}
    \label{fig:planbench_example}
\end{figure}

PlanBench (see Figure~\ref{fig:planbench_example}), as introduced by \citet{valmeekam2024planbench}, has been the most widely used benchmark for evaluating planning capabilities of LLMs. It provides a \emph{template} to convert symbolic model information into natural language text that can be used to either train or test LLMs on plan generation. An innovation of PlanBench is its ``obfuscated'' planning domains, where names are replaced using misleading vocabulary (see Appendix~\ref{app:obfuscated_prompts}). This aims to assess whether LLMs can plan based solely on the logical structure of the planning problem without any linguistic cues.

To better support both training and evaluating LLMs' planning skills, we extend the dataset in two key ways\footnote{The dataset will be released via Huggingface Hub.}:

\noindent\textbf{1. New Domains.}\quad We expand PlanBench to include additional domains from the IPC benchmarks, increasing the total from 2 to 8 domains. This allows for evaluating the model across diverse domains and complexities, providing a comprehensive assessment of its planning skills.

\noindent\textbf{2. Longer-plan Problems.}\quad A significant limitation observed in \emph{PlanGPT} \citep{rossetti2024learning} was that both the training and test data shared the same distribution, a limitation discussed in \S~\ref{sec:intro}. To address this, we have deliberately generated test instances with \emph{longer plan-lengths}, which is arguably the most straightforward way to generate OOD performance. By doing it, we prevent LLMs from easily retrieving familiar plan trace patterns from the training data, forcing them to engage in actual planning. 


The updated benchmark consists of: 

\noindent\textbf{1. Train Set:} 4000 instances per domain, with plan lengths from 3 to 16 actions. This contrasts with the previous training set, which consisted of 70,000 instances and had no restrictions on plan length.

\noindent\textbf{2. Test Sets}: 200 distinct instances per domain, all of which are not present in the training set, categorized into four groups: \textbf{(a) In-Distrib}: same plan length distribution as the training data. \textbf{(b) Long} (\emph{new category}): longer plan length distribution ranging from 17 to 32 actions. \textbf{(c) Unseen} (\emph{new category}): two novel domains not seen and not trained on by the model. \textbf{(d) Obfuscated}: two domains with obfuscated predicates, actions and objects.

To prevent the model from reusing memorized plan traces and achieving artificially high performance through pattern matching, we limit the training data to 4000 problem instances per domain. While the data size might still seem substantial, it is important to recognize that deep learning models, particularly modern LLMs, are data-hungry and are typically trained on trillions of tokens. This size represents only 5.7\% of the data used in the \emph{PlanGPT} work.

\begin{figure}[t]
    \centering
    \begin{tcolorbox}[title=Permutation Augmentation in Query Content, height=8.70cm, fontupper=\scriptsize\ttfamily\fontfamily{cmtt}\selectfont]
        \begin{minipage}[t]{0.45\textwidth}
            \begin{small}
                Original:
            \end{small}
            \\
            \highlight{gray30}{Query:}\\
            \dots\\
            \highlight{yellow20}{[ACTION DESCRIPTION]}\\
            Action a\\
            Action b\\
            Action c\\
            \dots\\
            Action a's precon 1\\
            Action a's precon 2\\
            Action a's effect 1\\
            Action a's effect 2\\
            \dots\\
            \highlight{yellow20}{[STATEMENT]}\\
            Initial state:\\
            1. orange block is clear\\
            2. hand is clear\\
            \dots\\
            My goal is that:\\
            1. red on white\\
            2. blue on orange\\
            \dots\\
        \end{minipage}
        \hfill
        \begin{minipage}[t]{0.5\textwidth}
            \begin{small}
                + Permutation:
            \end{small}
            \\
            \highlight{gray30}{Query:}\\
            \dots\\
            \highlight{yellow20}{[ACTION DESCRIPTION]}\\
            Action c \textcolor{red}{$\leftarrow$}\\
            Action a \textcolor{red}{$\leftarrow$}\\
            Action b \textcolor{red}{$\leftarrow$}\\
            \dots\\
            Action a's effect 2 \textcolor{blue}{$\leftarrow$}\\
            Action c's precon 1 \textcolor{blue}{$\leftarrow$}\\
            Action b's effect 1 \textcolor{blue}{$\leftarrow$}\\
            Action a's precon 2 \textcolor{blue}{$\leftarrow$}\\
            \dots\\
            \highlight{yellow20}{[STATEMENT]}\\
            Initial state:\\
            1. hand is clear \textcolor{ForestGreen}{$\leftarrow$}\\
            2. orange block is clear \textcolor{ForestGreen}{$\leftarrow$}\\
            \dots\\
            My goal is that:\\
            1. blue on orange \textcolor{ForestGreen}{$\leftarrow$}\\
            2. red on white \textcolor{ForestGreen}{$\leftarrow$}\\
            \dots\\
        \end{minipage}
    \end{tcolorbox}
    \caption{Permutation augmentation strategy shuffles the order of action descriptions (\textcolor{red}{red arrows}), condition and effect descriptions (\textcolor{blue}{blue arrows}), and atoms in initial and goal statements (\textcolor{ForestGreen}{green arrows}). The model is expected to learn underlying semantics through more diverse data representation, avoiding overfitting to superficial patterns.}
    \label{fig:planbench_permute}
\end{figure}

\subsection{Strategies for Enhancing Reasoning}
\label{sec:approach}
This section presents four strategies aiming to enhance LLMs' reasoning skills: \emph{permutation augmentation}, \emph{chain of thought}, \emph{self-correction}, and \emph{reinforcement learning}. We briefly explain the rationale and adaptation of each strategy to the planning domain.

\begin{figure}[t]
    \centering
    \begin{tcolorbox}[title=CoT Prompts to the Plan Response, height=8.70cm, fontupper=\scriptsize\ttfamily\fontfamily{cmtt}\selectfont]
        \begin{minipage}[t]{0.5\textwidth}
            \begin{small}
                + Goal CoT:
            \end{small}
            \\
            \highlight{gray30}{Response:}\\
        My plan is as follows:\\
        \highlight{yellow20}{[PLAN]}\\
        \highlight{magenta30}{[GOAL]}\\
        \tboxcomment{Repeat the goal}\\
        My goal is \dots \\
        \highlight{magenta30}{[GOAL END]}\\
        \highlight{magenta30}{[COUNT]}\\
        \tboxcomment{Count the remaining step to the goal}\\
        10 \\
        \highlight{magenta30}{[COUNT END]}\\
        put down the orange block\\
        \tboxcomment{Next step in the plan}\\
        \highlight{magenta30}{[GOAL]}\\
        My goal is \dots \\
        \highlight{magenta30}{[GOAL END]}\\
        \highlight{magenta30}{[COUNT]}\\
        9 \\
        \dots\\
        \highlight{yellow20}{[PLAN END]}
        \end{minipage}
        \hfill
        \begin{minipage}[t]{0.48\textwidth}
            \begin{small}
                + State CoT:
            \end{small}
            \\
            \highlight{gray30}{Response:}\\
        My plan is as follows:\\
        \highlight{yellow20}{[PLAN]}\\
        \highlight{magenta30}{[PRECON]}\\
        \tboxcomment{Provide grounded conditions for the action}\\
        I can only pick up a yellow block if the yellow block is on the table and the yellow block is clear. \\
        \dots\\
        \highlight{magenta30}{[PRECON END]}\\
        put down the orange block\\
        \highlight{magenta30}{[EFFECT]}\\
        \tboxcomment{Provide grounded effects}\\
        Once I put down a yellow block, my hand becomes empty.\\
        \highlight{magenta30}{[EFFECT END]}\\
        \dots\\
        \highlight{yellow20}{[PLAN END]}
        \end{minipage}
    \end{tcolorbox}
    \caption{Two types of CoT prompts are used in the plan response -- \emph{Goal CoT} and \emph{State CoT}. \emph{Goal CoT} prompts the agent to repeat the goal and count the remaining steps to the goal. \emph{State CoT} prompts the agent to provide grounded conditions for the action and grounded effects. The model is expected to learn the world dynamics through these prompts.}
    \label{fig:planbench_cot}
\end{figure}

\subsubsection{Permutation Augmentation}
\label{subsec:permute}
Early studies have shown that data augmentations enhancing input diversity can improve a model's generalization performance \citep{cubuk2019randaugment, gontijo2020affinity}. Recently, \citet{allen2023physics} discovered that permutation augmentation -- randomly rearranging sentences in a query -- significantly improved \emph{Transformer} models for question-answering tasks. They posit that exposing models to varied expressions of the same knowledge encourages them to discover the underlying structure rather than overfitting to superficial word patterns. We hypothesize a more specific rationale for why permutation augmentation works: it enhances the \emph{self-attention} mechanism in \emph{Transformer} during training. By permuting sentences, it ensures that when a token \emph{attends} to others, the attended tokens do not consistently occupy the same positions. This prevents the \emph{attention} mechanism from exploiting shortcuts based on positional or syntactic cues, which could potentially skew the model's semantic understanding. Instead, permutation encourages each token to engage with semantically relevant tokens, enhancing the quality of each token's embedding. These improved embeddings enhance the model's overall ability to predict the answers.

As such, we apply it to the planning instance descriptions in the training data. Specifically, we permute the order of action descriptions and statement sentences, as illustrated in Figure~\ref{fig:planbench_permute}. These permuted variants remain semantically equivalent, similar to how any permutation of predicates within conjunctions or disjunctions in PDDL expressions preserves their semantic meaning.

\subsubsection{Chain of Thought (CoT)}
\label{subsec:cot}

Producing intermediate steps before directly predicting the final output, is the core idea behind CoT prompting, a strategy that has shown promise in eliciting LLMs' reasoning, as demonstrated by numerous studies \citep{wang2023self, shi2023language, yao2024tree}. Complementing these empirical studies, \citet{PrystawskiLG23} provided a theoretical rationale for the effectiveness of chain-of-thought reasoning. Using a controlled Bayesian network environment, they showed that generating intermediate variables before predicting the target variable can improve a model's ability to match true conditional probabilities, particularly when relevant local variables are observed near each other in the network.

To design general CoT prompts for planning tasks, we draw inspiration from Decision Transformer \citep{chen2021decision}, a RL work that uses LMs as policy models for Atari Benchmark. Both RL and planning address sequential decision problems, sharing key elements such as initial states, state transition functions, and reward/heuristic functions. In RL, Decision Transformer conditions action prediction on the provided remaining \emph{return} and \emph{state} information. Recognizing this common ground, we structured our CoT prompts to also include the information of the goal distance and the current state. We posit that training LLMs to predict these details before deciding on an action would help the model learn world dynamics better and thus make better decisions.

But, in our context, serializing fully-observable states directly into natural language can lead to memory explosion, potentially overwhelming the model's capacity. Thus, we instead represent states using transition information (see Figure~\ref{fig:planbench_cot}). In fact, by estimating these details, LLMs perform the roles of both state transition and heuristic functions in classical search algorithms. An orthogonal investigation by \citet{katz2024thought} also explored leveraging LLMs to generate state transition and heuristic function code for planning tasks, but essentially not examining the plan generation skills.

\subsubsection{Self-Correction Learning}
\label{subsec:self_correction}

\begin{figure}[t]
    \centering
    \begin{tcolorbox}[title=Response with Erroneous Steps and Correction, height=5.70cm, fontupper=\scriptsize\ttfamily\fontfamily{cmtt}\selectfont]
            \begin{small}
                + Self Correction:
            \end{small}
            \\
            \highlight{gray30}{Response:}\\
        My plan is as follows:\\
        \tboxcomment{Assume the length of this reference plan is 7.}\\
        \highlight{yellow20}{[PLAN]}\\
        $1^{\text{st}}$ action; $2^{\text{nd}}$ action;\\
        $5^{\text{th}}$ action \highlight{purple30}{[WRONG]};  \tboxcomment{Synthesized mistakes by selecting an action that appears later in the plan}\\
        $3^{\text{rd}}$ action; $4^{\text{th}}$ action;\\
        $7^{\text{th}}$ action \highlight{purple30}{[WRONG]};   \tboxcomment{[WRONG] is special `removal' token}\\
        $5^{\text{th}}$ action;\\
        $6^{\text{th}}$ action; $7^{\text{th}}$ action\\
        \highlight{yellow20}{[PLAN END]}
        
    \end{tcolorbox}
    \caption{Self-correction learning modifies plan responses to involve erroneous steps followed by their immediate corrections, with the incorrect step indicated by token \texttt{[WRONG]}.}
    \label{fig:planbench_self_correct}
\end{figure}

This strategy has shown effectiveness especially in grade school math: \citet{kumar2024training} demonstrated that a multi-turn online reinforcement learning approach, which trains the model to correct its own mistakes, improves the model's accuracy. Similarly, \citet{Ye2024PhysicsOL} showed that training LLMs on data containing errors and their ``immediate removal'' can lead to higher accuracy compared to training on error-free data alone. Several theoretical frameworks support the effectiveness of self-correction learning, with the most intuitive being \emph{contrastive learning}. By exposing the model to both mistakes and correct solutions, we establish a robust learning environment. In contrast, training solely on ground truth data can lead to \emph{exposure bias}, where the model becomes overly confident in its predictions due to a limited range of correct examples. This overconfidence can hinder the model's ability to generalize effectively to unseen data \citep{an2022cont}.

This strategy has been largely overlooked in previous studies on training LLMs for plan generation. To explore its impact, we adopt an approach similar to \citet{Ye2024PhysicsOL}, designing synthesized error-correction procedure in the planning steps. Specifically, we randomly select an action that appears later in the plan sequence and insert it to the current step with a special `removal' token, as illustrated in Figure~\ref{fig:planbench_self_correct}.

\subsubsection{Reinforcement Learning (RL)}
\label{subsec:rl}
More recently, \citet{liu2024deepseek} found that when applying RL specifically to reasoning tasks, such as math and code generation, yields superior performance gains compared to applying RL to more general tasks. This suggests that RL could be a particularly well-suited strategy for enhancing LLM planning skills.

However, applying RL training for LLM-based plan generation poses two key challenges: (1) Typical RL frameworks for LLM are designed to attribute rewards to individual tokens as they are generated. However, in plan generation, the validity of a plan can only be determined once the entire response is complete, which creates a mismatch with the token-level optimization paradigm. (2) Using plan validity as the reward signal -- assigning $+1$ only when plans are valid, and $0$ rewards otherwise -- results in very sparse feedback. Such binary and infrequent rewards often lead to poor performance in RL algorithms \citep{ecoffet2019go}. 

To address the first challenge, we leverage the recently proposed \emph{rloo} framework by
\citet{ahmadian2024back}. \emph{rloo} enables LLMs to optimize at the sequence level, providing delayed rewards associated with the entire output rather than individual tokens. This approach aligns better with planning tasks, where plan validity can only be assessed upon completion of the entire sequence.

To address the second challenge and create a more informative reward signal, we propose using the \emph{Longest Contiguous Common Subsequence (LCCS)}. Let $P_g$ be the generated plan and $P_r$ be the reference plan. We define the LCCS reward function as:
\begin{align}
    R(P_g, P_r) = \begin{cases}
    1 & \text{if } P_g \text{ is valid} \\
    \frac{|\text{LCCS}(P_g, P_r)|}{|P_r|} & \text{otherwise}
    \end{cases}
\end{align}
When the plan is valid, we continue giving a reward of $+1$. However, when the plan is invalid, a supplementary reward is also provided based on the length of the LCCS between $P_g$ and $P_r$ (also see Figure~\ref{fig:next_token_prediction_illu} part b). Note that the current reward system has an inherent bias because it relies on a single reference plan, whereas there may be multiple valid plans for a given problem. However, this limitation can be effectively addressed in future work by considering distances to multiple reference plans from top-k planners, potentially providing a more robust measure. Despite this limitation, we found that \emph{LCCS} serves as a good reward signal to fill the sparsity gap and provide granular feedback on the quality of the output. As such, it offers smoother gradients for the model to learn from, facilitating more effective training.

\section{Evaluation Results}
\subsubsection*{Model}
We fine-tuned the open-source LLM \textsc{Qwen2-7B-Instruct} \citep{yang2024qwen2} on the extended PlanBench dataset. This model was pre-trained on general text but not on code. Note that \textsc{Qwen2}'s architecture shares significant similarities with \textsc{Llama 3} \citep{dubey2024llama}, but was trained on a smaller dataset. It thus reduce the risk of having exposed to PlanBench or related data during pre-training. See training configurations and other details in Appendix~\ref{app:implementation_details}.

\subsubsection*{Evaluation Metrics}
\label{subsec:metrics}
In previous work, the assessment of LLM planners primarily focused on the \emph{validity rate} of the generated plans \citep{kambhampatiposition2024, stechly2024chain}. However, the validity rate alone is inadequate for differentiating between the performances of various strategies, as it is too stringent for evaluating incremental improvements in plan quality. Relying solely on this metric can obscure deeper insights into the planning process, preventing a thorough understanding of where the model faces reasoning bottlenecks and how to address them. To address this limitation, we introduce a complementary metric: \emph{executability rate}. A plan is deemed \emph{executable} if it satisfies all preconditions for the actions within the current state. This metric allows for a finer evaluation of plan quality, even if the plan does not completely achieve the goal state. To prevent trivially empty plans from being considered executable, we only count plans with more than 3 actions in the evaluation.

In the following sections, we present our experimental results, focusing on the effectiveness of the four strategies for enhancing LLM planning capabilities. We evaluate performance in both in-distribution and out-of-distribution test sets to provide a comprehensive assessment. To begin with, we establish a baseline by fine-tuning the LLM on the vanilla corpus. During the training phase, the model is tasked with predicting the next tokens in serialized planning instances consisting of domain descriptions, problem instance descriptions, and the associated plans. Then, in the testing phase, we prompt the model with only the domain and problem instance descriptions. The model is expected to continue the sequence by generating the plan\footnote{Code available at \url{https://anonymous.4open.science/r/official-misconcept-lm-plan-gen-D34B}}.

\begin{table}[t]
    \centering
    \caption{Performance of the fine-tuned LLM across various test sets with no additional strategies applied. Although the LLM performs well on in-distribution, it struggles to generalize to OOD cases.}
    \label{tab:baseline}
    \resizebox{0.9\columnwidth}{!}{%
    \begin{tabular}{@{}l|cccc@{}}
    \toprule
    \multirow{2}{*}{\textbf{Domain}} & \multicolumn{2}{c|}{\textbf{In-Distrib.}}                        & \multicolumn{2}{c}{\textbf{Long}}         \\ \cmidrule(l){2-5} 
                            & \emph{valid. rate} & \multicolumn{1}{c|}{\emph{exec. rate}}    & \emph{valid. rate}  & \emph{exec. rate} \\ \midrule
    \textsc{blocksworld}             & 98.5\%            & \multicolumn{1}{c|}{98.5\%}         & \badval 13.5\%                    & \badval 23.5\%      \\
    \textsc{logistics}               & \textbf{100\%}    & \multicolumn{1}{c|}{\textbf{100\%}} & \badval 14.0\%                    & \badval 20.5\%      \\
    \textsc{barman}                  & \textbf{100\%}    & \multicolumn{1}{c|}{\textbf{100\%}} & 25.0\%                    & 43.5\%      \\
    \textsc{childsnack}              & \textbf{100\%}    & \multicolumn{1}{c|}{\textbf{100\%}} & \goodval 66.0\%          & \goodval 67.0\%      \\
    \textsc{depots}                  & 98.5\%            & \multicolumn{1}{c|}{98.5\%}         & 31.0\%                    & 37.0\%      \\
    \textsc{driverlog}               & \textbf{100\%}    & \multicolumn{1}{c|}{\textbf{100\%}} & 31.0\%                    & 50.7\%      \\
    \textsc{grippers}                & 99.0\%            & \multicolumn{1}{c|}{99.0\%}         & 50.5\%                    & \goodval 76.0\%      \\
    \textsc{satellite}               & 99.0\%            & \multicolumn{1}{c|}{99.2\%}         & 51.5\%                    & 53.0\%      \\ \midrule
    \multirow{2}{*}{\textbf{Domain}} & \multicolumn{4}{c}{\textbf{Unseen}}                                                                 \\ \cmidrule(l){2-5} 
                            & \multicolumn{2}{c}{\emph{valid. rate}}                   & \multicolumn{2}{c}{\emph{exec. rate}}  \\ \midrule
    \textsc{hanoi}                   & \multicolumn{2}{c}{\badval   \textbf{0\%}              }                        & \multicolumn{2}{c}{   35\%              } \\
    \textsc{storage}                 & \multicolumn{2}{c}{\badval   \textbf{0\%}              }                        & \multicolumn{2}{c}{\badval   \textbf{1\%}              } \\ \midrule
    \multirow{2}{*}{\textbf{Domain}} & \multicolumn{4}{c}{\textbf{Obfuscated}}                                                             \\ \cmidrule(l){2-5} 
                            & \multicolumn{2}{c}{\emph{valid. rate}}                   & \multicolumn{2}{c}{\emph{exec. rate}}  \\ \midrule
    \textsc{blocksworld}             & \multicolumn{2}{c}{\badval   \textbf{0\%}              }                        & \multicolumn{2}{c}{\badval   \textbf{0\%}              } \\
    \textsc{logistics}               & \multicolumn{2}{c}{\badval   \textbf{0\%}              }                        & \multicolumn{2}{c}{\badval   \textbf{0\%}              } \\ \bottomrule
    \end{tabular}%
    }
    \end{table}

\subsection{LLMs Learn to Plan in Natural Language, but Struggle in OOD Scenarios}
\label{sec:result_1}
Previous studies have asserted the effectiveness of fine-tuning LLMs for plan generation \citep{rossetti2024learning, shah2024causal}. We revisit this claim, examining whether the statements hold true in our extended PlanBench dataset. 

\subsubsection{In-Distribution Test: A Promising Start} Table~\ref{tab:baseline} presents the performance of our fine-tuned LLM on the vanilla corpus. The model indeed achieved high performance across all domains in in-distribution tests, mirroring the positive result reported by \emph{PlanGPT} \citep{rossetti2024learning}. Remarkably, we attain comparable performance using only 5.7\% of their training data and a more complex input format expressed in natural language, suggesting not only better data efficiency but also better capability to process less structured data than previously anticipated.

\subsubsection{Longer Planning: A Drastic Decline}
The `long' test set reveals a significant performance drop, particularly in the NP-hard \textsc{Blocksworld} domain \citep{Chenoweth1991OnTN}, where the \emph{validity rate} falls from 98.5\% to 13.5\%. This dramatic decline underscores the model's limitations in handling longer and more complex planning scenarios, suggesting that the planning capabilities acquired through end-to-end fine-tuning are not robust.

\subsubsection{\textsc{Childsnack}: Partial Success?}

In the \textsc{Childsnack} domain, which involves preparing sandwiches for allergic and not allergic children, the model achieves the highest validity rate at 66.0\% in the long' test set. However, the reasoning complexity of the tasks in this domain is not very sensitive to plan length. While the plan length expands with an increase in the number of children to feed, the order in which the children are addressed does not matter. Upon examining the generated plans, it appears that the model's major shortcoming is \textbf{not} its ability to handle food allergies, which it manages adeptly. Instead, it is the model's failure to adapt to updated world dynamics involving more children and objects, often resulting in the omission of several children from the plan. Therefore, it suggests that the model's reasoning abilities are insufficient for adapting to changes in out-of-distribution scenarios, even when the core task logic is well comprehended.

\begin{figure*}[t]
    \begin{minipage}{0.75\linewidth}
        \captionof{table}{Ablation Study on Strategy Effectiveness in Planning. Validity rates (valid.) and the Executability rate (exec.) are analyzed. Strategies such as Permutation, CoT, and Self-Correct show no significant \emph{validity}. improvements but enhance \emph{executability} in `long' and other OOD scenarios. Notably, `Goal CoT' appears to hinder performance. We attribute this to the dual duties of generating plans and accurately estimating the heuristics of the state, which increases overall complexity and hinders the model's ability to effectively learn both aspects. RL emerges as the only strategy that enhances \emph{validity} in OOD scenarios. Improvements of statistical significance are highlighted in \textcolor{ForestGreen}{green}, while significant declines are highlighted in \textcolor{red}{red}.}
    \label{tab:ablation_study}
    \centering
    \resizebox{0.9\textwidth}{!}{%
    \begin{tabular}{@{}c|ccccc|cc|cc|cc|cc@{}}
    \toprule
                                     & \multicolumn{5}{c|}{\textbf{Strategies}}                                                                                                                                                                         & \multicolumn{2}{c|}{\textbf{In-Distrib.}} & \multicolumn{2}{c|}{\textbf{Long}}                              & \multicolumn{2}{c|}{\textbf{Unseen}}          & \multicolumn{2}{c}{\textbf{Obfuscated}} \\ \cmidrule(l){2-14} 
    \multirow{-2}{*}{\textbf{Label}} & Perm.                           & \begin{tabular}[c]{@{}c@{}}Goal\\ CoT\end{tabular} & \begin{tabular}[c]{@{}c@{}}State\\ CoT\end{tabular} & \begin{tabular}[c]{@{}c@{}}Self\\ Correct\end{tabular} & RL         & valid.            & exec.                    & valid.                           & exec.                           & valid.                           & exec.         & valid.               & exec.               \\ \midrule
    1                                & \multicolumn{1}{c|}{}           & \multicolumn{1}{c|}{}                              & \multicolumn{1}{c|}{}                               & \multicolumn{1}{c|}{}                                  &            & 99.3\%          & 99.8\%                  & 34.8\%                         & 42.3\%                           & \textbf{0\%}                   & 20.1\%                              & \textbf{0\%}       & \textbf{0\%}                               \\ \midrule
    2                                & \multicolumn{1}{c|}{\checkmark} & \multicolumn{1}{c|}{}                              & \multicolumn{1}{c|}{}                               & \multicolumn{1}{c|}{}                                  &            & 99.5\%          & 99.8\%                  & 35.0\%                         & \goodval                48.3\%   & \textbf{0\%}                   & \goodval                75.5\%      & \textbf{0\%}       & \textbf{0\%}                               \\ \midrule
    3                                & \multicolumn{1}{c|}{\checkmark} & \multicolumn{1}{c|}{\checkmark}                    & \multicolumn{1}{c|}{}                               & \multicolumn{1}{c|}{}                                  &            & 96.8\%          & 98.5\%                  & \badval                 12.1\% & \badval 18.7\%                   & \goodval                5.5\%  & 53.4\%                              & \textbf{0\%}       & \textbf{0\%}                               \\ \midrule
    4                                & \multicolumn{1}{c|}{\checkmark} & \multicolumn{1}{c|}{}                              & \multicolumn{1}{c|}{\checkmark}                     & \multicolumn{1}{c|}{}                                  &            & 98.9\%          & 99.5\%                  & 29.5\%                         &                          43.0\%  & \textbf{0\%}                   & \goodval \textbf{100\%}             & \textbf{0\%}       & \textbf{0\%}                               \\ \midrule
    5                                & \multicolumn{1}{c|}{\checkmark} & \multicolumn{1}{c|}{\checkmark}                    & \multicolumn{1}{c|}{\checkmark}                     & \multicolumn{1}{c|}{}                                  &            & 98.7\%          & 99.0\%                  & \badval                 23.8\% & 39.3\%                           & \textbf{0\%}                   & \goodval 90.8\%                     & \textbf{0\%}       & \textbf{0\%}                               \\ \midrule
    6                                & \multicolumn{1}{c|}{\checkmark} & \multicolumn{1}{c|}{}                              & \multicolumn{1}{c|}{}                               & \multicolumn{1}{c|}{\checkmark}                        &            & 99.7\%          & 99.9\%                  & 32.6\%                         & \goodval 50.6\%                  & \textbf{0\%}                   & \goodval 70.9\%                     & \textbf{0\%}       & \textbf{0\%}                               \\ \midrule
    7                                & \multicolumn{1}{c|}{\checkmark} & \multicolumn{1}{c|}{\checkmark}                    & \multicolumn{1}{c|}{}                               & \multicolumn{1}{c|}{\checkmark}                        &            & 97.3\%          & 98.6\%                  & \badval                 14.9\% & \badval 25.6\%                   & \textbf{0\%}                   & 38.7\%                              & \textbf{0\%}       & \textbf{0\%}                               \\ \midrule
    8                                & \multicolumn{1}{c|}{\checkmark} & \multicolumn{1}{c|}{}                              & \multicolumn{1}{c|}{\checkmark}                     & \multicolumn{1}{c|}{\checkmark}                        &            & 98.1\%          & 99.3\%                  & 27.5\%                         & \goodval                 49.4\%  & \textbf{0\%}                   & \goodval 94.5\%                     & \textbf{0\%}       & \textbf{0\%}                               \\ \midrule
    9                                & \multicolumn{1}{c|}{\checkmark} & \multicolumn{1}{c|}{\checkmark}                    & \multicolumn{1}{c|}{\checkmark}                     & \multicolumn{1}{c|}{\checkmark}                        &            & 98.3\%          & 99.1\%                  & \badval                 25.9\% & \badval 30.4\%                   & \textbf{0\%}                   & \goodval 90.6\%                     & \textbf{0\%}       & \textbf{0\%}                               \\ \midrule
    \textbf{10$\star$}               & \multicolumn{1}{c|}{}           & \multicolumn{1}{c|}{}                              & \multicolumn{1}{c|}{}                               & \multicolumn{1}{c|}{}                                  & \checkmark & 99.2\%          & 99.6\%                  & \goodval                41.5\% & \goodval                 51.3\%  & \goodval                12.5\% & \badval 23.0\%                      & \textbf{0\%}       & \textbf{0\%}                               \\ \midrule
    \textbf{11$\star$}               & \multicolumn{1}{c|}{}           & \multicolumn{1}{c|}{}                              & \multicolumn{1}{c|}{}                               & \multicolumn{1}{c|}{\checkmark}                        & \checkmark & 99.7\%          & \textbf{100\%}          & 36.3\%                         & \goodval                 53.6\%  & \textbf{0\%}                   & \goodval 71.5\%                     & \textbf{0\%}       & \textbf{0\%}                               \\ \bottomrule
    \end{tabular}%
    }
    \end{minipage}%
    \hfill
    \begin{minipage}{0.23\linewidth}
        \begin{minipage}{0.95\linewidth}  
            \centering
            \captionof{table}{Probing test: LLM effectively identifies when it makes mistakes.}
            \resizebox{0.65\textwidth}{!}{%
                \begin{tabular}{@{}c|cc@{}}
                \toprule
                \textbf{Row} & \textbf{Precision} & \textbf{Recall} \\ \midrule
                6                & 87.5               & 98.1            \\ \midrule
                7                & 89.3               & 99.2            \\ \midrule
                8                & 89.2               & 97.4            \\ \midrule
                \textbf{9$\star$}& 90.5               & 99.2            \\ \bottomrule
                \end{tabular}%
                }
            \label{tab:mistake_identification}
        \end{minipage}
        
        \vspace{0.1cm}
        
        \begin{minipage}{0.8\linewidth}  
            \centering
            \includegraphics[width=\textwidth]{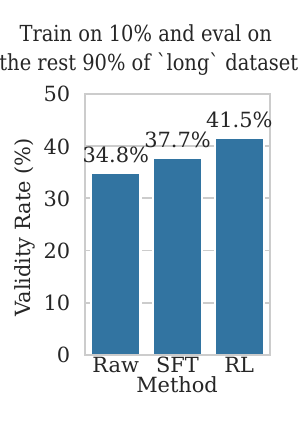}
        \end{minipage}
        \vspace{-0.7cm}
        \caption{RL achieves higher gain than SFT in `long' test.}
        \label{fig:sft_rl_success_rate_comparison}
    \end{minipage}
\end{figure*}

\subsubsection{Generalization to Novel Domains: A Clear Failure}

The fine-tuned model utterly failed to perform in the ``unseen'' and ``obfuscated'' test sets, unable to generate either valid or executable plans. This performance breakdown was particularly striking in the ``obfuscated'' test set. Here, the LLM planner often resorted to repeating irrelevant actions from domains present in the training set, neglecting the actual planning context (see Appendix~\ref{app:results_examples} for more details).

Overall, our results refute the claim that fine-tuning alone enables LLMs to master complex planning. Next, we will examine whether the purported strategies can turn the tide.

\subsection{The Secret Help of Permutation}
For the sake of brevity, Table~\ref{tab:ablation_study} presents only the average performance across all domains in the ablation study. Our analysis reveals an intriguing finding regarding the impact of permutation augmentation. While this technique does not significantly improve the \emph{validity rate}, it largely enhances the \emph{executability rate} (see Table~\ref{tab:ablation_study} row 2). In particular, we observe a remarkable 75.5\% score in ``unseen'' test set, while the vanilla model only got 20.1\% (row 1). This high performance suggests that permutation augmentation enables the model to effectively parse unseen problem content, which includes unseen actions, predicates and objects. This aligns with the findings of \citet{allen2023physics} and underscores the importance of data augmentation in enhancing LLMs capacity to interpret and adapt to new contexts.

\subsection{Goal CoT: The Complexity Paradox and Overfitting Issue}

\emph{Goal CoT} is the only strategy that hinders planning performance among OOD cases, showing no improvement whatsoever (see Table~\ref{tab:ablation_study} row 3, 5, 7, 9). We attribute the failure to two factors:
\noindent\textbf{1. Complexity Paradox:} Estimating the goal distance adds complexity to the planning process. Although the intention was to provide heuristic guidance, this added layer paradoxically complicates the task. The requirement to predict the steps needed to achieve a goal demands precision but also restricts the model's flexibility during planning -- By fixing the total number of action steps before planning begins, the model loses the ability to dynamically adjust its plan based on the evolving conditions.
\noindent\textbf{2. Poor Generalization:} The model exhibits a noticeable bias towards estimating numbers within the range of plan lengths that it has previously encountered during the training stage. This aligns with the observable limitations of LLMs in generalizing to unseen formulas and number -- an issue that has been extensively highlighted by \citet{gorceix24learning}.

\subsection{LLMs Recognize Mistakes But Fail to Correct Them}

Despite high initial expectations for self-correction learning -- stemming from its demonstrated effectiveness in solving maths, this recently proposed strategy did not improve \emph{validity rates} (see Table~\ref{tab:ablation_study} row 6, 7, 8, 9). To understand this outcome, we conducted \emph{mistake identification probing tests} on test sets to assess the conditional probability of outputting the special `removal' token when the model encounters a wrong step (see Appendix~\ref{app:mistake_identification_probing_tests} for more details). 

Results from Table~\ref{tab:mistake_identification} showed that the model is able to accurately identify errors, achieving particularly high precision (90.5\%) and recall (99.2\%) when all 4 strategies are combined (row 9). However, the detection capability does not lead to effective correction, indicating that future research should focus on how to leverage detected errors for correction. Nevertheless, the strategy gave a slight improvement in \emph{executability rate} (e.g., row 10 vs. 11), suggesting its ability to enhance the \emph{coherence} of generated sequences. 

\subsection{State CoT Boosts Executability with a Caveat: Efficacy Limited to Short Problems}
\label{sec:result_state_cot_not_ok}

We observed that \emph{State CoT} does not improve plan \emph{executability} within the `long' test set, yet it significantly enhances performance within the `unseen' test set (e.g., 100\% in row 4). Importantly, the `unseen' test set retains the same plan length distribution as the training set. Thus, we posit that the \emph{State CoT}'s ability to enhance the model's understanding of state transition dynamics may likely be limited to the plan length distribution it encountered during training. Consequently, we do not observe an improvement in the `long' test set. This also rationalizes why the \emph{State CoT} demonstrates efficacy in other reasoning tasks \citep{wei2022chain, yao2024tree}, where these tasks typically require solution steps that align with the training data distribution. We shall verify this hypothesis in the next section through a `plan continuation' experiment.

\subsection{Familiar-Length Plan Continuation Experiments Reveal CoT's Potential}

A critical question arises regarding the model's poor performance on longer problems within seen domains: Is this drop caused by distribution shift? Given that the model was trained on short-plan problem, it could have developed a bias towards shorter plans. To investigate this, we conducted a `plan continuation' experiment, where we provided the first 15 true actions and asked the model to continue from there, ensuring that the remaining steps fell within the length distribution seen during training.

\begin{figure}
    \centering
    \includegraphics[width=0.7\columnwidth]{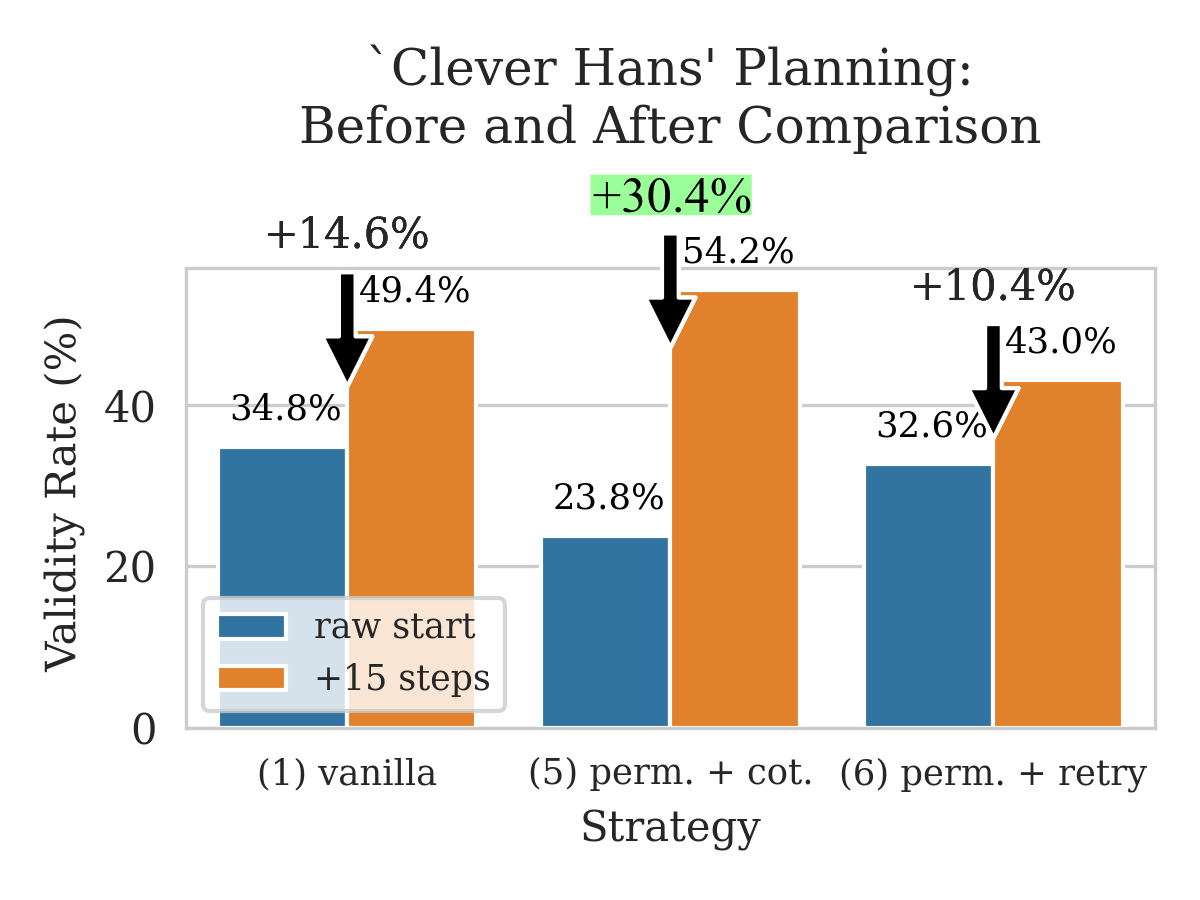}
    \vspace{-0.55cm}
    \caption{Validity rate of the LLM planner on the `long' test set when provided with the first 15 actions. The model using CoT (Goal + State) showed the highest performance gain when hints were provided.}
    \label{fig:clever_hans_results}
\end{figure}

The results, as shown in Figure~\ref{fig:clever_hans_results}, reveal intriguing insights. Despite the significant hint provided, the model's performance still lags behind that of the in-distribution test set. This discrepancy is expected, as the model must infer the current world state from the given actions and then continue planning to reach the goal state. Even with the initial actions provided, predicting the world state remains challenging.

Interestingly, the model employing CoT (Goal + State) demonstrates the highest performance gain when provided with the hints. Its validity rate improves dramatically from the lowest (23.8\%) to the highest (54.2\%) among the compared strategies. This finding suggests that when CoT operates within its ``comfort zone'' (i.e., in-distribution scenarios), it begins to show its effectiveness in enhancing the model's planning, supporting the hypothesis presented in \S~\ref{sec:result_state_cot_not_ok}. While this performance boost is encouraging, it also highlights a limitation: CoT appears to be overfit to in-distribution inference. This aligns with our earlier observation that the model faces difficulty estimating the goal distance that is not within the training distribution.

\subsection{RL Enhances Model Performance}

\begin{figure}
    \centering
    \includegraphics[width=0.70\columnwidth]{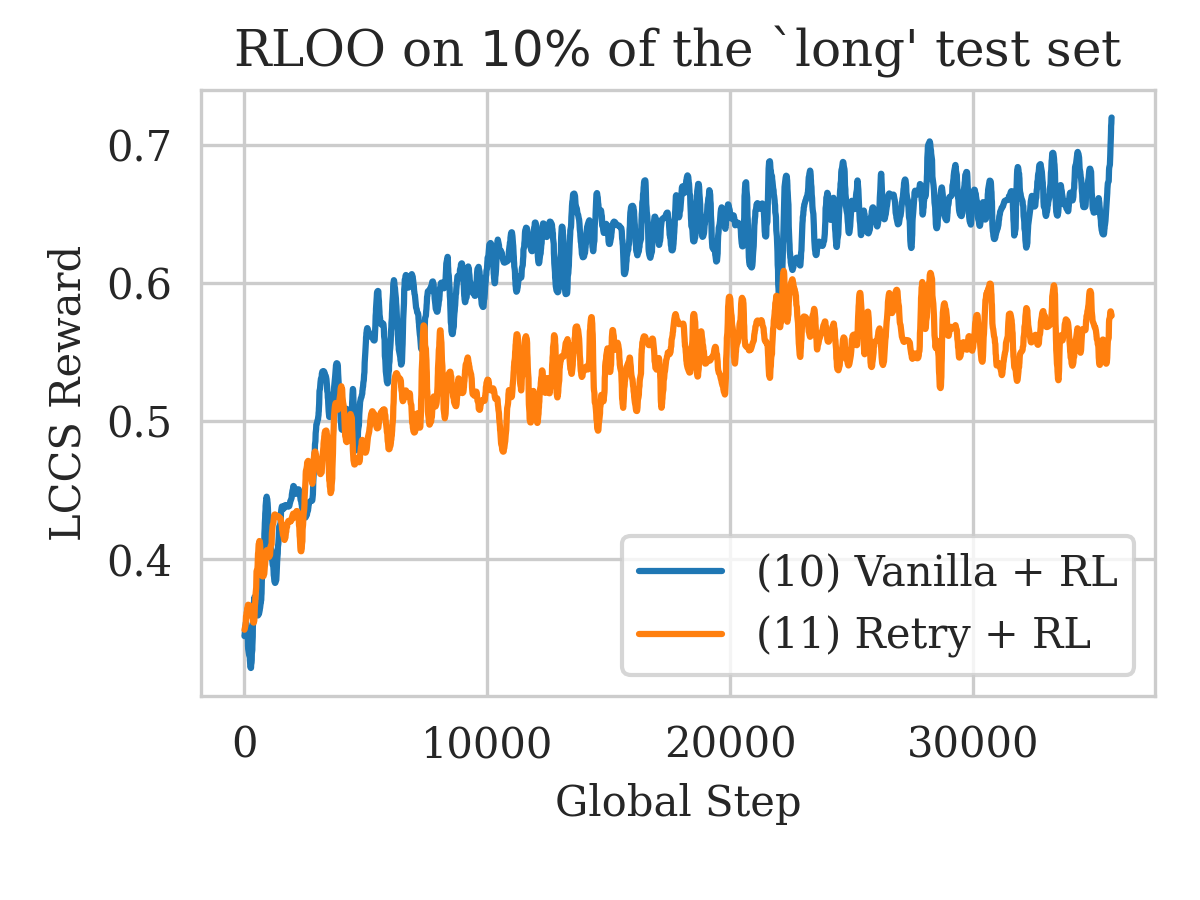}
    \vspace{-0.6cm}
    \caption{Reward curve of the RL training process. The LLM get further trained on the 10\% of the `long' test set with limited training steps. This however lead to a noticeable performance gain in OOD cases.}
    \label{fig:rloo_reward_lineplot}
\end{figure}

RL notably improves the performance under our end-to-end planning paradigm, especially on longer problems. Note that the model was trained on 10\% of the `long' test set with the proposed \emph{LCCS}-based reward model, and evaluated on the 90\% of the `long' test set and other OOD test sets.

Despite the limited training data and suboptimal rewards achieved on this subset, RL boosted the validity rate on the `long' test set from 34.8\% to 41.5\% (a 6.7\% increase) and the executability rate from 42.3\% to 53.6\% (9.0\%) (see Table \ref{tab:ablation_study}, row 10).  Interestingly, it also enabled the model to solve problems in the `unseen' test set, achieving a 12.5\% where it previously failed to generate any valid plans. The updated model does not exhibit overfitting, as indicated by the \emph{LCCS} reward signal not reaching a perfect score of 1.0. Instead, the model has developed general planning strategies effective in unseen scenarios. To confirm that this improvement is not due to the additional training data, we also conducted supervised fine-tuning as described in Section~\ref{sec:result_1} using the same training data. However, the outcomes were not as promising as those achieved with RL, as demonstrated in Figure~\ref{fig:sft_rl_success_rate_comparison}. These results suggest that RL fosters more comprehensive planning skills compared to supervised fine-tuning (SFT), aligning with the findings of \citet{liu2024deepseek}.

Applying RL to the vanilla model led to faster convergence and improved results compared to its application to the model with self-correction skills, as illustrated in Figure~\ref{fig:rloo_reward_lineplot} and row 10 and 11 in Table~\ref{tab:ablation_study}. We hypothesize that the self-correction strategy, by permitting repeated attempts at actions, effectively broadens the model's state space and thus poses a greater challenge to explore a valid solution.

\section{Discussion}

We investigate the enigma of why strategies aimed at improving LLM reasoning often fail to achieve expected results in planning tasks. Our findings reveal that, they do contribute to the overall plan quality, as reflected in the enhanced \emph{executability rate}. This indicates progress towards more \textbf{coherent} plans, despite not directly leading to valid plans. We have limited our scope to the end-to-end plan generation paradigm, where the \emph{search process} is not explicitly programmed. Within this context, we find that fine-tuning LLMs on datasets consisting solely of problems and corresponding reference plans struggles to foster robust planning skills beyond in-distribution instances. Nonetheless, our research reveals that RL stands out as the most effective strategy in this end-to-end paradigm, enhancing both the \emph{validity} and \emph{executability} rates on longer problems. This study provides a clear pathway for boosting the planning capability within the \emph{next-token prediction} framework. Whereas previous work centered on pinpointing the shortcoming of LLMs in plan generation, we show that the \textbf{direction} of improving LLM planning lies in increasing the likelihood of reaching goal state while preserving the high levels of executability that current strategies have already achieved.

\bibliography{aaai25}

\begin{thebibliography}{39}
\providecommand{\natexlab}[1]{#1}

\bibitem[{Ahmadian et~al.(2024)Ahmadian, Cremer, Gall{\'e}, Fadaee, Kreutzer, {\"U}st{\"u}n, and Hooker}]{ahmadian2024back}
Ahmadian, A.; Cremer, C.; Gall{\'e}, M.; Fadaee, M.; Kreutzer, J.; {\"U}st{\"u}n, A.; and Hooker, S. 2024.
\newblock Back to basics: Revisiting reinforce style optimization for learning from human feedback in llms.
\newblock \emph{arXiv preprint arXiv:2402.14740}.

\bibitem[{Allen-Zhu and Li(2023)}]{allen2023physics}
Allen-Zhu, Z.; and Li, Y. 2023.
\newblock Physics of language models: Part 3.1, knowledge storage and extraction.
\newblock \emph{arXiv preprint arXiv:2309.14316}.

\bibitem[{An et~al.(2022)An, Feng, Lv, Kong, Qiu, and Huang}]{an2022cont}
An, C.; Feng, J.; Lv, K.; Kong, L.; Qiu, X.; and Huang, X. 2022.
\newblock Cont: Contrastive neural text generation.
\newblock \emph{Advances in Neural Information Processing Systems}, 35: 2197--2210.

\bibitem[{Bachmann et~al.(2024)}]{bachmann2024pitfalls}
Bachmann, G.; et~al. 2024.
\newblock The pitfalls of next-token prediction.
\newblock \emph{arXiv preprint arXiv:2403.06963}.

\bibitem[{Cao et~al.(2024)Cao, Katz, Kokel, Srinivas, and Sohrabi}]{cao2024automating}
Cao, D.; Katz, M.; Kokel, H.; Srinivas, K.; and Sohrabi, S. 2024.
\newblock Automating Thought of Search: A Journey Towards Soundness and Completeness.
\newblock \emph{arXiv preprint arXiv:2408.11326}.

\bibitem[{Chen et~al.(2020)Chen, Ding, Edwards, Chau, Hou, Johnson, Sharukh~Syed, Tang, Wu, Yan, Gil, and Nir}]{chen2020planimation}
Chen, G.; Ding, Y.; Edwards, H.; Chau, C.~H.; Hou, S.; Johnson, G.; Sharukh~Syed, M.; Tang, H.; Wu, Y.; Yan, Y.; Gil, T.; and Nir, L. 2020.
\newblock Planimation.
\newblock \emph{arXiv preprint arXiv:2008.04600}.

\bibitem[{Chen et~al.(2024)Chen, Liao, Li, and Fan}]{chen2024alphamath}
Chen, G.; Liao, M.; Li, C.; and Fan, K. 2024.
\newblock AlphaMath Almost Zero: Process Supervision without Process.
\newblock In \emph{Proceedings of the Neural Information Processing Systems (NeurIPS)}.

\bibitem[{Chen et~al.(2021)Chen, Lu, Rajeswaran, Lee, Grover, Laskin, Abbeel, Srinivas, and Mordatch}]{chen2021decision}
Chen, L.; Lu, K.; Rajeswaran, A.; Lee, K.; Grover, A.; Laskin, M.; Abbeel, P.; Srinivas, A.; and Mordatch, I. 2021.
\newblock Decision transformer: Reinforcement learning via sequence modeling.
\newblock \emph{Advances in neural information processing systems}, 34: 15084--15097.

\bibitem[{Chenoweth(1991)}]{Chenoweth1991OnTN}
Chenoweth, S.~V. 1991.
\newblock On the NP-Hardness of Blocks World.
\newblock In \emph{AAAI Conference on Artificial Intelligence}.

\bibitem[{Cubuk et~al.(2019)Cubuk, Zoph, Shlens, and Le}]{cubuk2019randaugment}
Cubuk, E.~D.; Zoph, B.; Shlens, J.; and Le, Q.~V. 2019.
\newblock Randaugment: Practical data augmentation with no separate search.
\newblock \emph{arXiv preprint arXiv:1909.13719}, 2(4): 7.

\bibitem[{Dubey et~al.(2024)Dubey, Jauhri, Pandey, Kadian, Al-Dahle, Letman, Mathur, Schelten, Yang, Fan et~al.}]{dubey2024llama}
Dubey, A.; Jauhri, A.; Pandey, A.; Kadian, A.; Al-Dahle, A.; Letman, A.; Mathur, A.; Schelten, A.; Yang, A.; Fan, A.; et~al. 2024.
\newblock The llama 3 herd of models.
\newblock \emph{arXiv preprint arXiv:2407.21783}.

\bibitem[{Ecoffet et~al.(2019)Ecoffet, Huizinga, Lehman, Stanley, and Clune}]{ecoffet2019go}
Ecoffet, A.; Huizinga, J.; Lehman, J.; Stanley, K.~O.; and Clune, J. 2019.
\newblock Go-explore: a new approach for hard-exploration problems.
\newblock \emph{arXiv preprint arXiv:1901.10995}.

\bibitem[{Gontijo-Lopes et~al.(2020)Gontijo-Lopes, Smullin, Cubuk, and Dyer}]{gontijo2020affinity}
Gontijo-Lopes, R.; Smullin, S.~J.; Cubuk, E.~D.; and Dyer, E. 2020.
\newblock Affinity and diversity: Quantifying mechanisms of data augmentation.
\newblock \emph{arXiv preprint arXiv:2002.08973}.

\bibitem[{Gorceix et~al.(2024)Gorceix, Le~Chenadec, Rammal, Vadori, and Veloso}]{gorceix24learning}
Gorceix, A.; Le~Chenadec, B.; Rammal, A.; Vadori, N.; and Veloso, M. 2024.
\newblock Learning Mathematical Rules with Large Language Models.
\newblock In \emph{The 4th Workshop on Mathematical Reasoning and AI at NeurIPS'24}.

\bibitem[{Haslum et~al.(2019)Haslum, Lipovetzky, Magazzeni, Muise, Brachman, Rossi, and Stone}]{haslum2019introduction}
Haslum, P.; Lipovetzky, N.; Magazzeni, D.; Muise, C.; Brachman, R.; Rossi, F.; and Stone, P. 2019.
\newblock \emph{An introduction to the planning domain definition language}, volume~13.
\newblock Springer.

\bibitem[{Howey et~al.(2004)}]{HoweyLF04}
Howey, R.; et~al. 2004.
\newblock {VAL:} Automatic Plan Validation, Continuous Effects and Mixed Initiative Planning Using {PDDL}.
\newblock In \emph{{ICTAI}}, 294--301. {IEEE} Computer Society.

\bibitem[{Huang et~al.(2023)Huang, Yu, Ma, Zhong, Feng, Wang, Chen, Peng, Feng, Qin, and Liu}]{Huang2023ASO}
Huang, L.; Yu, W.; Ma, W.; Zhong, W.; Feng, Z.; Wang, H.; Chen, Q.; Peng, W.; Feng, X.; Qin, B.; and Liu, T. 2023.
\newblock A Survey on Hallucination in Large Language Models: Principles, Taxonomy, Challenges, and Open Questions.
\newblock \emph{ArXiv}, abs/2311.05232.

\bibitem[{Kambhampati et~al.(2024)Kambhampati, Valmeekam, Guan, Verma, Stechly, Bhambri, Saldyt, and Murthy}]{kambhampatiposition2024}
Kambhampati, S.; Valmeekam, K.; Guan, L.; Verma, M.; Stechly, K.; Bhambri, S.; Saldyt, L.~P.; and Murthy, A.~B. 2024.
\newblock Position: LLMs Can't Plan, But Can Help Planning in LLM-Modulo Frameworks.
\newblock In \emph{Forty-first International Conference on Machine Learning}.

\bibitem[{Katz et~al.(2024)Katz, Kokel, Srinivas, and Sohrabi}]{katz2024thought}
Katz, M.; Kokel, H.; Srinivas, K.; and Sohrabi, S. 2024.
\newblock Thought of Search: Planning with Language Models Through The Lens of Efficiency.
\newblock In \emph{The First Workshop on System-2 Reasoning at Scale, NeurIPS'24}.

\bibitem[{Kumar et~al.(2024)Kumar, Zhuang, Agarwal, Su, Co-Reyes, Singh, Baumli, Iqbal, Bishop, Roelofs et~al.}]{kumar2024training}
Kumar, A.; Zhuang, V.; Agarwal, R.; Su, Y.; Co-Reyes, J.~D.; Singh, A.; Baumli, K.; Iqbal, S.; Bishop, C.; Roelofs, R.; et~al. 2024.
\newblock Training Language Models to Self-Correct via Reinforcement Learning.
\newblock \emph{arXiv preprint arXiv:2409.12917}.

\bibitem[{Liu et~al.(2024)Liu, Feng, Wang, Wang, Liu, Zhao, Dengr, Ruan, Dai, Guo et~al.}]{liu2024deepseek}
Liu, A.; Feng, B.; Wang, B.; Wang, B.; Liu, B.; Zhao, C.; Dengr, C.; Ruan, C.; Dai, D.; Guo, D.; et~al. 2024.
\newblock Deepseek-v2: A strong, economical, and efficient mixture-of-experts language model.
\newblock \emph{arXiv preprint arXiv:2405.04434}.

\bibitem[{Liu et~al.(2023)Liu, Bubeck, Eldan, Kulkarni, Li, Nguyen, Ward, and Zhang}]{liu2023tinygsm}
Liu, B.; Bubeck, S.; Eldan, R.; Kulkarni, J.; Li, Y.; Nguyen, A.; Ward, R.; and Zhang, Y. 2023.
\newblock Tinygsm: achieving> 80\% on gsm8k with small language models.
\newblock \emph{arXiv preprint arXiv:2312.09241}.

\bibitem[{Madaan et~al.(2024)Madaan, Tandon, Gupta, Hallinan, Gao, Wiegreffe, Alon, Dziri, Prabhumoye, Yang et~al.}]{madaan2024self}
Madaan, A.; Tandon, N.; Gupta, P.; Hallinan, S.; Gao, L.; Wiegreffe, S.; Alon, U.; Dziri, N.; Prabhumoye, S.; Yang, Y.; et~al. 2024.
\newblock Self-refine: Iterative refinement with self-feedback.
\newblock \emph{Advances in Neural Information Processing Systems}, 36.

\bibitem[{Mirzadeh et~al.(2024)Mirzadeh, Alizadeh, Shahrokhi, Tuzel, Bengio, and Farajtabar}]{mirzadeh2024gsm}
Mirzadeh, I.; Alizadeh, K.; Shahrokhi, H.; Tuzel, O.; Bengio, S.; and Farajtabar, M. 2024.
\newblock GSM-Symbolic: Understanding the Limitations of Mathematical Reasoning in Large Language Models.
\newblock \emph{arXiv preprint arXiv:2410.05229}.

\bibitem[{Prystawski, Li, and Goodman(2023)}]{PrystawskiLG23}
Prystawski, B.; Li, M.; and Goodman, N.~D. 2023.
\newblock Why think step by step? Reasoning emerges from the locality of experience.
\newblock In \emph{NeurIPS}.

\bibitem[{Rossetti et~al.(2024)Rossetti, Tummolo, Gerevini, Putelli, Serina, Chiari, and Olivato}]{rossetti2024learning}
Rossetti, N.; Tummolo, M.; Gerevini, A.~E.; Putelli, L.; Serina, I.; Chiari, M.; and Olivato, M. 2024.
\newblock Learning General Policies for Planning through GPT Models.
\newblock In \emph{Proceedings of the International Conference on Automated Planning and Scheduling}, volume~34, 500--508.

\bibitem[{Shah et~al.(2024)Shah, Dikkala, Wang, and Panigrahy}]{shah2024causal}
Shah, K.; Dikkala, N.; Wang, X.; and Panigrahy, R. 2024.
\newblock Causal Language Modeling Can Elicit Search and Reasoning Capabilities on Logic Puzzles.
\newblock \emph{arXiv preprint arXiv:2409.10502}.

\bibitem[{Shi et~al.(2023)Shi, Suzgun, Freitag, Wang, Srivats, Vosoughi, Chung, Tay, Ruder, Zhou, Das, and Wei}]{shi2023language}
Shi, F.; Suzgun, M.; Freitag, M.; Wang, X.; Srivats, S.; Vosoughi, S.; Chung, H.~W.; Tay, Y.; Ruder, S.; Zhou, D.; Das, D.; and Wei, J. 2023.
\newblock Language models are multilingual chain-of-thought reasoners.
\newblock In \emph{{ICLR}}. OpenReview.net.

\bibitem[{Stechly et~al.(2024)}]{stechly2024chain}
Stechly, K.; et~al. 2024.
\newblock Chain of thoughtlessness: An analysis of CoT in planning.
\newblock \emph{arXiv preprint arXiv:2405.04776}.

\bibitem[{Valmeekam et~al.(2024{\natexlab{a}})Valmeekam, Marquez, Olmo, Sreedharan, and Kambhampati}]{valmeekam2024planbench}
Valmeekam, K.; Marquez, M.; Olmo, A.; Sreedharan, S.; and Kambhampati, S. 2024{\natexlab{a}}.
\newblock Planbench: An extensible benchmark for evaluating large language models on planning and reasoning about change.
\newblock \emph{Advances in Neural Information Processing Systems}, 36.

\bibitem[{Valmeekam et~al.(2023)Valmeekam, Marquez, Sreedharan, and Kambhampati}]{valmeekam2023planning}
Valmeekam, K.; Marquez, M.; Sreedharan, S.; and Kambhampati, S. 2023.
\newblock On the planning abilities of large language models-a critical investigation.
\newblock \emph{Advances in Neural Information Processing Systems}, 36: 75993--76005.

\bibitem[{Valmeekam et~al.(2022)Valmeekam, Olmo, Sreedharan, and Kambhampati}]{Valmeekam2022PlanBenchAE}
Valmeekam, K.; Olmo, A.; Sreedharan, S.; and Kambhampati, S. 2022.
\newblock PlanBench: An Extensible Benchmark for Evaluating Large Language Models on Planning and Reasoning about Change.
\newblock In \emph{Neural Information Processing Systems}.

\bibitem[{Valmeekam et~al.(2024{\natexlab{b}})Valmeekam, Stechly, Gundawar, and Kambhampati}]{valmeekam2024planning}
Valmeekam, K.; Stechly, K.; Gundawar, A.; and Kambhampati, S. 2024{\natexlab{b}}.
\newblock Planning in Strawberry Fields: Evaluating and Improving the Planning and Scheduling Capabilities of LRM o1.
\newblock \emph{arXiv preprint arXiv:2410.02162}.

\bibitem[{Wang et~al.(2023{\natexlab{a}})Wang, Wei, Schuurmans, Le, Chi, Narang, Chowdhery, and Zhou}]{wang2023self}
Wang, X.; Wei, J.; Schuurmans, D.; Le, Q.~V.; Chi, E.~H.; Narang, S.; Chowdhery, A.; and Zhou, D. 2023{\natexlab{a}}.
\newblock Self-Consistency Improves Chain of Thought Reasoning in Language Models.
\newblock In \emph{{ICLR}}. OpenReview.net.

\bibitem[{Wang et~al.(2023{\natexlab{b}})Wang, Wei, Schuurmans, Le, Chi, Narang, Chowdhery, and Zhou}]{2WSLCNCZ23}
Wang, X.; Wei, J.; Schuurmans, D.; Le, Q.~V.; Chi, E.~H.; Narang, S.; Chowdhery, A.; and Zhou, D. 2023{\natexlab{b}}.
\newblock Self-Consistency Improves Chain of Thought Reasoning in Language Models.
\newblock In \emph{{ICLR}}. OpenReview.net.

\bibitem[{Wei et~al.(2022)Wei, Wang, Schuurmans, Bosma, Xia, Chi, Le, Zhou et~al.}]{wei2022chain}
Wei, J.; Wang, X.; Schuurmans, D.; Bosma, M.; Xia, F.; Chi, E.; Le, Q.~V.; Zhou, D.; et~al. 2022.
\newblock Chain-of-thought prompting elicits reasoning in large language models.
\newblock \emph{Advances in neural information processing systems}, 35: 24824--24837.

\bibitem[{Yang et~al.(2024)Yang, Yang, Hui, Zheng, Yu, Zhou, Li, Li, Liu, Huang et~al.}]{yang2024qwen2}
Yang, A.; Yang, B.; Hui, B.; Zheng, B.; Yu, B.; Zhou, C.; Li, C.; Li, C.; Liu, D.; Huang, F.; et~al. 2024.
\newblock Qwen2 technical report.
\newblock \emph{arXiv preprint arXiv:2407.10671}.

\bibitem[{Yao et~al.(2024)Yao, Yu, Zhao, Shafran, Griffiths, Cao, and Narasimhan}]{yao2024tree}
Yao, S.; Yu, D.; Zhao, J.; Shafran, I.; Griffiths, T.; Cao, Y.; and Narasimhan, K. 2024.
\newblock Tree of thoughts: Deliberate problem solving with large language models.
\newblock \emph{Advances in Neural Information Processing Systems}, 36.

\bibitem[{Ye et~al.(2024)Ye, Xu, Li, and Allen-Zhu}]{Ye2024PhysicsOL}
Ye, T.; Xu, Z.; Li, Y.; and Allen-Zhu, Z. 2024.
\newblock Physics of Language Models: Part 2.2, How to Learn From Mistakes on Grade-School Math Problems.
\newblock \emph{ArXiv}, abs/2408.16293.

\end{thebibliography}

\onecolumn
\appendix
The appendix consists of the following:
\begin{itemize}
    \item Appendix~\ref{app:terminology}: Terminology Explanation: Executability and Validity of a Plan
    \item Appendix~\ref{app:implementation_details}: Implementation Details
    \item Appendix~\ref{app:results_examples}: Planning Prompts and Responses
    \item Appendix~\ref{app:self_correction_learning_details}: Further Details on A Failure Case of Self-Correction Learning
    \item Appendix~\ref{app:mistake_identification_probing_tests}: Further Details on Mistake Identification Probing Tests
    \item Appendix~\ref{app:pass_at_k}: Additional Results: \emph{pass@k} Validity Rate
    \item Appendix~\ref{app:goal_satisfiability_result}: Additional Results: Goal Satisfiability Rate
\end{itemize}

Our codebase is available at \url{https://anonymous.4open.science/r/official-misconcept-lm-plan-gen-D34B} and the dataset will be released via Huggingface Hub once the review process is completed.

\section{Terminology Explanation: Executability and Validity of a Plan}
\label{app:terminology}
\addcontentsline{toc}{section}{entry}

The definition below can be found in \citet{HoweyLF04} and is also presented as \textsc{Definition 2.8} in the textbook \citep{haslum2019introduction}, defined as follows:

\begin{definition}[Executability of a Plan]
    A plan is executable if it defines an action sequence $(a_0, a_1, \ldots, a_{n-1})$ with states $(s_0, s_1, \ldots, s_n)$. $s_0$ is the initial state and for each $i=0, 1, \ldots, n-1$, $s_{i+1}$ is the result of executing $a_i$ in $s_i$, and the precondition of $a_i$ must hold in $s_i$ and $s_{i+1}$ is the result of removing delete effects, adding add effects and applying numeric effects. The state $s_n$ is called the final state produced by the plan and the state sequence $(s_0, s_1, \ldots, s_n)$ is called the trace of the plan. An executable plan produce a unique trace. 
\end{definition}

\begin{definition}[Validity of a Plan]
    A plan is valid if and only if it is executable and the goal of the problem $G$ holds in the final state $s_n$ produced by the plan.
\end{definition}

These are the formal definitions written in the PDDL textbook, and we can see that executability is a prerequisite for a plan to be valid. This underscores the importance of including the executability metric in our evaluation of plan quality.

\section{Implementation Details}
\label{app:implementation_details}

We used 4 Nvidia A100 GPUs to fine-tune the \textsc{Qwen2-7B-Instruct} model on the extended PlanBench dataset. Note that during the training process, especially during reinforcement learning, the machine occasionally encounter out of memory (OOM) errors, even when reducing the batch size to 1. This issue primarily arose because when applying Chain-of-Thought (CoT) prompts to the model, the length of the response sequences often exceeded 20,000 tokens, which is beyond what most open-source frameworks can handle.

Below are the detailed hyperparameter used in our experiments.

\subsection{Hyperparameter}
\label{app:hyperparameters}

\begin{table}[H]
    \centering
    \resizebox{0.7\columnwidth}{!}{%
    \begin{tabular}{@{}lllp{6cm}@{}}
    \toprule
    Topic                            & Hyperparameter              & Value                  & Notes                                                                                                                                                           \\ \midrule
    \multirow{4}{*}{Data generation} & mistake rate                & 0.2                    & Follow the best practice in paper                                                                                                                               \\
                                     & seed                        & 1111                   &                                                                                                                                                                 \\
                                     & train size                  & 4000                   &                                                                                                                                                                 \\
                                     & test size                   & 200                    &                                                                                                                                                                 \\ \midrule
    \multirow{9}{*}{Fine-tuning}     & model name                  & Qwen2-7B-Instruct      &                                                                                                                                                                 \\
                                     & stage                       & sft                    &                                                                                                                                                                 \\
                                     & deepspeed                   & zero 3                 &                                                                                                                                                                 \\
                                     & per device train batch size & 1 - 4                  & Adaptive to the response length                                                                                                                                 \\
                                     & gradient accumulation steps & 1 - 2                  &                                                                                                                                                                 \\
                                     & num train epochs            & 2.0                    &                                                                                                                                                                 \\
                                     & bf16                        & true                   &                                                                                                                                                                 \\
                                     & lr scheduler type           & cosine                 &                                                                                                                                                                 \\
                                     & learning rate               & 1.0e-5                 &                                                                                                                                                                 \\ \midrule
    \multirow{12}{*}{RL}             & lora rank                   & 128                    &                                                                                                                                                                 \\
                                     & lora alpha                  & 256                    &                                                                                                                                                                 \\
                                     & per device train batch size & 1                      &                                                                                                                                                                 \\
                                     & gradient accumulation steps & 2-4                    &                                                                                                                                                                 \\
                                     & if qlora                    & false                  &                                                                                                                                                                 \\
                                     & seed                        & 23                     &                                                                                                                                                                 \\
                                     & num ppo epochs              & 1                      &                                                                                                                                                                 \\
                                     & num mini batches            & 1                      &                                                                                                                                                                 \\
                                     & learning rate               & 3.0e-6                 &                                                                                                                                                                 \\
                                     & total episodes              & 500000                 &                                                                                                                                                                 \\
                                     & missing eos penalty         & 0.5                    &                                                                                                                                                                 \\
                                     & lora dropout                & 0.05                   &                                                                                                                                                                 \\ \midrule
    \multirow{3}{*}{Evaluation}      & k                           & 1, 3, 5                & Number of plan candidates to generate                                                                                                                           \\
                                     & top p                       & 0.93                   & Top P Sampling coefficient when k $>$ 1, if P is set to 0.75, the model will include enough top tokens so that their combined probability equals or exceeds 75\%. \\
                                     & top k                       & 50                     & If it is set to 50, only the 50 most likely tokens are considered for the next output token                                                                     \\ \midrule
    \end{tabular}%
    }
    \end{table}

\section{Problem Instances Prompts and Responses}
\label{app:results_examples}

\subsection{Detailed Examples of Training Corpus from \textsc{Driverlog} Domain}

We will release the extended PlanBench dataset via Huggingface Hub after anonymous review. In this section, we provide several examples of problem instance context and responses text from the \textsc{Driverlog} domain.

To illustrate modifications to the response section of the training corpus, we use row 9 from Table~\ref{tab:ablation_study} as an example, where all four strategies are combined.

Figure~\ref{fig:problem_context_example} will display the problem context section of the example. Following that, Figure~\ref{fig:response_example_detail} presents a detailed response section that incorporates the \emph{Goal CoT}, \emph{State CoT}, and \emph{Self-Correction} strategies.

\begin{figure}[H]
    \centering
    \begin{tcolorbox}[title=Training Data Driverlog Query Content Example, height=16.6cm, fontupper=\scriptsize\ttfamily\fontfamily{cmtt}\selectfont, fontlower=\scriptsize\ttfamily\fontfamily{cmtt}\selectfont]
        \highlight{gray30}{Query:}\\
        \highlight{yellow20}{[CONTEXT]}\\
        I have to plan how to transport objects between locations using trucks driven by drivers. The drivers can walk between locations, board and disembark from trucks, and drive trucks between locations. The trucks can be loaded and unloaded with objects.\\
        \highlight{yellow20}{[ACTION DESCRIPTION]}\\
        \emph{Here are the actions that can be performed:}\\
        Load an object into a truck.\\
        Unload an object from a truck.\\
        A driver boards a truck.\\
        A driver disembarks from a truck.\\
        A driver drives a truck from one location to another location.\\A driver walks from one location to another location.\\

        \emph{The following are the restrictions on the actions:}\\
        An object can be loaded into a truck only if the object is at the same location as the truck.\\Once an object is loaded into a truck, the object is not at the location but is in the truck.\\An object can be unloaded from a truck only if the object is in the truck.\\Once an object is unloaded from a truck, the object is not in the truck but is at the location of the truck.\\A driver can board a truck only if the driver is at the same location as the truck.\\A driver can board a truck only if the truck is empty.\\Once a driver boards a truck, we say the driver is driving the truck.\\Once a driver boards a truck, the driver is no longer at the location.\\Once a driver boards a truck, the truck is no longer empty.\\A driver can disembark from a truck only if the driver is driving the truck.\\Once a driver disembarks from a truck, we say the driver is no longer driving the truck.\\Once a driver disembarks from a truck, the driver is at the location of the truck.\\Once a driver disembarks from a truck, the truck is empty.\\A driver can drive a truck from one location to another location only if the driver is driving the truck and the truck is at the from-location.\\A driver can drive a truck from one location to another location only if the from-location and to-location are linked.\\Once a driver drives a truck from one location to another location, the truck is at the to-location and is no longer at the from-location.\\A driver can walk from one location to another location only if the driver is at the from-location and the from-location and to-location are linked.\\Once a driver walks from one location to another location, the driver is at the to-location and is no longer at the from-location.\\

        \highlight{yellow20}{[STATEMENT]}\\
        As initial conditions I have that, place 1 and place 0 are linked, driver 2 is at place 1, path 1-0 and place 0 are connected, path 1-0 and place 1 are connected, truck 1 is empty, place 1 and path 1-0 are connected, driver 1 is at place 1, package 1 is at place 1, truck 1 is at place 1, place 0 and place 1 are linked, place 0 and path 1-0 are connected.\\

        My goal is to have that truck 1 is at place 1, driver 2 is at place 0, package 1 is at place 0.
    \end{tcolorbox}
    \caption{The LLM planner is trained on these lengthy natural language queries that describe the context, action description, initial and goal states in the Driverlog domain. It's important to recognize that the next-token prediction training of LLMs during fine-tuning isn't limited to predicting just the response tokens. In fact, the model is trained to predict all tokens, including those in the query context, similar to the pretraining process. This aspect is often overlooked in the planning community, where people may assume that the model is only trained to generate the plan response.}
    \label{fig:problem_context_example}
\end{figure}

\begin{figure}[H]
    \centering
    \begin{tcolorbox}[title=Training Data Response Section in Driverlog Domain When Applied CoT and Self-Correction, height=16.20cm, fontupper=\scriptsize\ttfamily\fontfamily{cmtt}\selectfont]
     
        \highlight{gray30}{Response:}\\
        My plan is as follows:\\
        \highlight{yellow20}{[PLAN]}\\
        \highlight{magenta30}{<GOAL>} My goal is to have that truck 1 is at place 1, driver 2 is at place 0, package 1 is at place 0. \highlight{magenta30}{<COUNT>} 6 \highlight{magenta30}{</COUNT>}  \highlight{magenta30}{<PRECON>} A driver 1 can board a truck 1 only if the driver 1 is at the same place 1 as the truck 1. A driver 1 can board a truck 1 only if the truck 1 is empty. \highlight{magenta30}{</PRECON>} driver 1 boards truck 1 at place 1 \highlight{magenta30}{<EFFECT>} Once a driver 1 boards a truck 1, the truck 1 is no longer empty. Once a driver 1 boards a truck 1, we say the driver 1 is driving the truck 1. Once a driver 1 boards a truck 1, the driver 1 is no longer at the place 1. \highlight{magenta30}{</EFFECT>} \\
        \highlight{magenta30}{<GOAL>} My goal is to have that truck 1 is at place 1, driver 2 is at place 0, package 1 is at place 0. \highlight{magenta30}{<COUNT>} 5 \highlight{magenta30}{</COUNT>}  \highlight{magenta30}{<PRECON>} An package 1 can be unloaded from a truck 1 only if the package 1 is in the truck 1. \highlight{magenta30}{</PRECON>} unload package 1 from truck 1 at place 0 \highlight{magenta30}{<EFFECT>} Once an package 1 is unloaded from a truck 1, the package 1 is not in the truck 1 but is at the place 0 of the truck 1. \highlight{magenta30}{</EFFECT>} \highlight{purple30}{WRONG} \\
        \highlight{magenta30}{<GOAL>} My goal is to have that truck 1 is at place 1, driver 2 is at place 0, package 1 is at place 0. \highlight{magenta30}{<COUNT>} 5 \highlight{magenta30}{</COUNT>}  \highlight{magenta30}{<PRECON>} An package 1 can be loaded into a truck 1 only if the package 1 is at the same place 1 as the truck 1. \highlight{magenta30}{</PRECON>} load package 1 into truck 1 at place 1 \highlight{magenta30}{<EFFECT>} Once an package 1 is loaded into a truck 1, the package 1 is not at the place 1 but is in the truck 1. \highlight{magenta30}{</EFFECT>} \\
        \highlight{magenta30}{<GOAL>} My goal is to have that truck 1 is at place 1, driver 2 is at place 0, package 1 is at place 0. \highlight{magenta30}{<COUNT>} 4 \highlight{magenta30}{</COUNT>}  \highlight{magenta30}{<PRECON>} A driver 2 can walk from place 1 to path 1-0 only if the driver 2 is at the place 1 and the place 1 and path 1-0 are linked. \highlight{magenta30}{</PRECON>} driver 2 walks from place 1 to path 1-0 \highlight{magenta30}{<EFFECT>} Once a driver 2 walks from place 1 to path 1-0, the driver 2 is at the path 1-0 and is no longer at the place 1. \highlight{magenta30}{</EFFECT>} \\
        \highlight{magenta30}{<GOAL>} My goal is to have that truck 1 is at place 1, driver 2 is at place 0, package 1 is at place 0.  \highlight{magenta30}{<COUNT>} 1 \highlight{magenta30}{</COUNT>}  \highlight{magenta30}{<PRECON>} A driver 2 can walk from path 1-0 to place 0 only if the driver 2 is at the path 1-0 and the path 1-0 and place 0 are linked. \highlight{magenta30}{</PRECON>} driver 2 walks from path 1-0 to place 0 \highlight{magenta30}{<EFFECT>} Once a driver 2 walks from path 1-0 to place 0, the driver 2 is at the place 0 and is no longer at the path 1-0. \highlight{magenta30}{</EFFECT>} \highlight{purple30}{WRONG} \\
        \highlight{magenta30}{<GOAL>} My goal is to have that truck 1 is at place 1, driver 2 is at place 0, package 1 is at place 0. \highlight{magenta30}{<COUNT>} 3 \highlight{magenta30}{</COUNT>}  \highlight{magenta30}{<PRECON>} A driver 2 can walk from path 1-0 to place 0 only if the driver 2 is at the path 1-0 and the path 1-0 and place 0 are linked. \highlight{magenta30}{</PRECON>} driver 2 walks from path 1-0 to place 0 \highlight{magenta30}{<EFFECT>} Once a driver 2 walks from path 1-0 to place 0, the driver 2 is at the place 0 and is no longer at the path 1-0. \highlight{magenta30}{</EFFECT>} \\
        \highlight{magenta30}{<GOAL>} My goal is to have that truck 1 is at place 1, driver 2 is at place 0, package 1 is at place 0. \highlight{magenta30}{<COUNT>} 2 \highlight{magenta30}{</COUNT>}  \highlight{magenta30}{<PRECON>} A driver 1 can drive a truck 1 from place 1 to place 0 only if the driver 1 is driving the truck 1 and the truck 1 is at the place 1. A driver 1 can drive a truck 1 from place 1 to place 0 only if the place 1 and place 0 are linked. \highlight{magenta30}{</PRECON>} driver 1 drives truck 1 from place 1 to place 0 \highlight{magenta30}{<EFFECT>} Once a driver 1 drives a truck 1 from place 1 to place 0, the truck 1 is at the place 0 and is no longer at the place 1. \highlight{magenta30}{</EFFECT>} \\
        \highlight{magenta30}{<GOAL>} My goal is to have that truck 1 is at place 1, driver 2 is at place 0, package 1 is at place 0. \highlight{magenta30}{<COUNT>} 1 \highlight{magenta30}{</COUNT>}  \highlight{magenta30}{<PRECON>} An package 1 can be unloaded from a truck 1 only if the package 1 is in the truck 1. \highlight{magenta30}{</PRECON>} unload package 1 from truck 1 at place 0 \highlight{magenta30}{<EFFECT>} Once an package 1 is unloaded from a truck 1, the package 1 is not in the truck 1 but is at the place 0 of the truck 1. \highlight{magenta30}{</EFFECT>} \\
        \highlight{magenta30}{<GOAL>} My goal is to have that truck 1 is at place 1, driver 2 is at place 0, package 1 is at place 0. \highlight{magenta30}{<COUNT>} 0 \highlight{magenta30}{</COUNT>}  \highlight{magenta30}{<PRECON>} A driver 1 can drive a truck 1 from place 0 to place 1 only if the driver 1 is driving the truck 1 and the truck 1 is at the place 0. A driver 1 can drive a truck 1 from place 0 to place 1 only if the place 0 and place 1 are linked. \highlight{magenta30}{</PRECON>} driver 1 drives truck 1 from place 0 to place 1 \highlight{magenta30}{<EFFECT>} Once a driver 1 drives a truck 1 from place 0 to place 1, the truck 1 is at the place 1 and is no longer at the place 0. \highlight{magenta30}{</EFFECT>}\\
        \highlight{yellow20}{[PLAN END]}
        
    \end{tcolorbox}
    \caption{The \emph{Goal CoT} contains two components, one the the $<$GOAL$>$ tag that prompts the agent to repeat the goal, and the other is the $<$COUNT$>$ tag that prompts the agent to count the remaining steps to the goal. The \emph{State CoT} contains two components, one is the $<$PRECON$>$ tag that prompts the agent to provide grounded conditions for the action, and the other is the $<$EFFECT$>$ tag that prompts the agent to provide grounded effects. Both the preconditions and effects are specifically tied to the actual objects present in the problem instance. In this example, there are two incorrect steps in the plan, which are marked as \emph{WRONG}. The first incorrect step borrowed the second to last step from the reference plan. The second step provides an incorrect value for the remaining steps to the goal.}
    \label{fig:response_example_detail}
\end{figure}

\subsubsection{Additinoal Detail for the Error Correction Data Synthesization}
In our current setting, the erroneous action step is generated by randomly selecting an action from later in the reference plan sequence and inserting it to the current position, followed by the \texttt{[WRONG]} token, indicating that the current step is incorrect and should be retried. This method introduces an inductive bias that encourages the model not to skip steps. That is, we want the model to not taking shortcuts to the states that are closer to the goal state, but to strictly obey the state transition dynamics. In our experiments, we also explored an alternative approach that just select irrelevant actions from the action space. This alternative approach will generate the erroneous step by modifying the current correct action step based on two strategies: (1) either keeping the action constant while changing to other valid objects or (2) keeping the object constant while changing to other valid actions. However, this alternative method produced worse results than our current approach. It suggests that the inductive bias introduced by our current method is beneficial for the model to learn the state transition dynamics.

\subsection{Obfuscated Prompts: \textsc{Blocksworld}}
\label{app:obfuscated_prompts}

Obfuscated domain replaces the name of the predicates, actions and objects with unfamiliar vocabulary. The following Figure~\ref{fig:app_obfuscated_prompt} is a comparison of the original and obfuscated domain descriptions for the \textsc{Blocksworld} domain:

\begin{figure}[H]
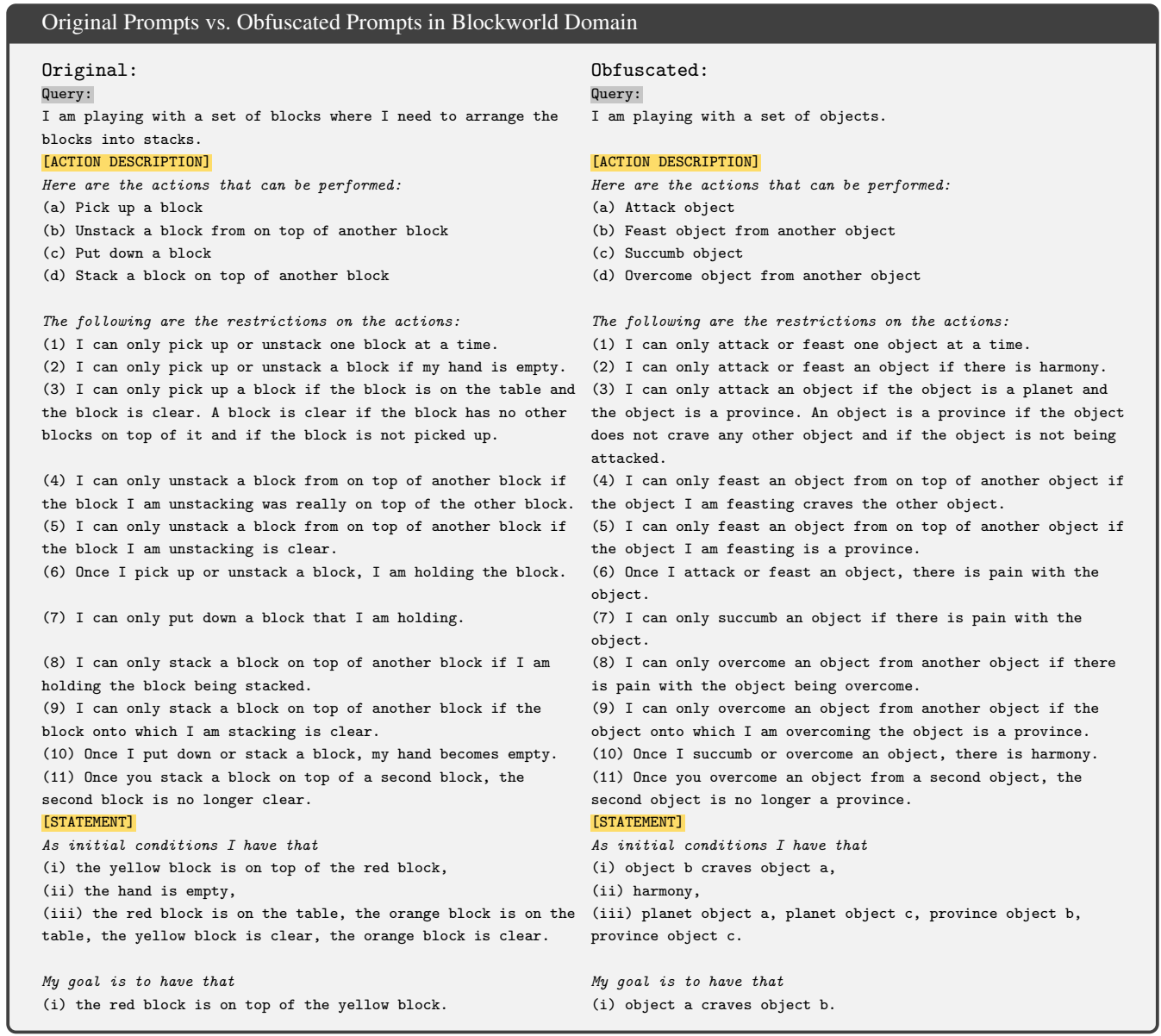

    \centering
    \begin{tcolorbox}[title=Original Prompts vs. Obfuscated Prompts in Blockworld Domain, height=15.90cm, fontupper=\scriptsize\ttfamily\fontfamily{cmtt}\selectfont]
        \begin{minipage}[t]{0.5\textwidth}
            \begin{small}
                Original:
            \end{small}
            \\
            \highlight{gray30}{Query:}\\
            I am playing with a set of blocks where I need to arrange the blocks into stacks.\\
            \highlight{yellow20}{[ACTION DESCRIPTION]}\\
            \emph{Here are the actions that can be performed:}\\
            (a) Pick up a block\\
            (b) Unstack a block from on top of another block\\
            (c) Put down a block\\
            (d) Stack a block on top of another block\\
            
            \emph{The following are the restrictions on the actions:}\\
            (1) I can only pick up or unstack one block at a time.\\
            (2) I can only pick up or unstack a block if my hand is empty.\\
            (3) I can only pick up a block if the block is on the table and the block is clear. A block is clear if the block has no other blocks on top of it and if the block is not picked up.\\

            (4) I can only unstack a block from on top of another block if the block I am unstacking was really on top of the other block.\\
            (5) I can only unstack a block from on top of another block if the block I am unstacking is clear.\\
            (6) Once I pick up or unstack a block, I am holding the block.\\

            (7) I can only put down a block that I am holding.\\
            
            (8) I can only stack a block on top of another block if I am holding the block being stacked.\\
            (9) I can only stack a block on top of another block if the block onto which I am stacking is clear.\\
            (10) Once I put down or stack a block, my hand becomes empty.\\
            (11) Once you stack a block on top of a second block, the second block is no longer clear.\\
            \highlight{yellow20}{[STATEMENT]}\\
            \emph{As initial conditions I have that}\\
            (i) the yellow block is on top of the red block,\\
            (ii) the hand is empty,\\
            (iii) the red block is on the table, the orange block is on the table, the yellow block is clear, the orange block is clear.\\

            \emph{My goal is to have that}\\
            (i) the red block is on top of the yellow block.\\
        \end{minipage}
        \hfill
        \begin{minipage}[t]{0.5\textwidth}
            \begin{small}
                Obfuscated:
            \end{small}
            \\
            \highlight{gray30}{Query:}\\
            I am playing with a set of objects.\\

            \highlight{yellow20}{[ACTION DESCRIPTION]}\\
            \emph{Here are the actions that can be performed:}\\
            (a) Attack object\\
            (b) Feast object from another object\\
            (c) Succumb object\\
            (d) Overcome object from another object\\
            
            \emph{The following are the restrictions on the actions:}\\
            (1) I can only attack or feast one object at a time.\\
            (2) I can only attack or feast an object if there is harmony.\\
            (3) I can only attack an object if the object is a planet and the object is a province. An object is a province if the object does not crave any other object and if the object is not being attacked.\\
            (4) I can only feast an object from on top of another object if the object I am feasting craves the other object.\\
            (5) I can only feast an object from on top of another object if the object I am feasting is a province.\\
            (6) Once I attack or feast an object, there is pain with the object.\\
            (7) I can only succumb an object if there is pain with the object.\\
            (8) I can only overcome an object from another object if there is pain with the object being overcome.\\
            (9) I can only overcome an object from another object if the object onto which I am overcoming the object is a province.\\
            (10) Once I succumb or overcome an object, there is harmony.\\
            (11) Once you overcome an object from a second object, the second object is no longer a province.\\
            \highlight{yellow20}{[STATEMENT]}\\
            \emph{As initial conditions I have that}\\
            (i) object b craves object a,\\
            (ii) harmony,\\
            (iii) planet object a, planet object c, province object b, province object c.\\

            \emph{My goal is to have that}\\
            (i) object a craves object b.\\
        \end{minipage}
    \end{tcolorbox}
    \caption{Comparison of original and obfuscated prompts in the blockworld domain. Obfuscated vocabularies for actions are: \emph{pick-up} $\rightarrow$ \emph{attack}, \emph{unstack} $\rightarrow$ \emph{feast}, \emph{put-down} $\rightarrow$ \emph{succumb}, \emph{stack} $\rightarrow$ \emph{overcome}; Obfuscated vocabularies for objects are: \emph{block} $\rightarrow$ \emph{object}; Obfuscated vocabularies for predicates are: \emph{on table} $\rightarrow$ \emph{planet}, \emph{clear} $\rightarrow$ \emph{province}, \emph{hand empty} $\rightarrow$ \emph{harmony}, \emph{holding} $\rightarrow$ \emph{pain}, \emph{on top of} $\rightarrow$ \emph{craves}.}
    \label{fig:app_obfuscated_prompt}
\end{figure}

The LLM planner had no exposure to the obfuscated domain during training. Our results showed that achieved 0\% validity and executability rates on the obfuscated test set. Upon examining the generated plans, we discovered an intriguing pattern: the model tends to ignore the obfuscated domain context and instead produced actions from the original Blocksworld domain. This behavior persisted even when we provided the first action step as a hint, as illustrated in Figure~\ref{fig:app_obfuscated_response}. 

\begin{figure}[H]
    \centering
    \begin{tcolorbox}[title=Response Example for Obfuscated Blockworld Problem, height=13.00cm, width=10cm, fontupper=\scriptsize\ttfamily\fontfamily{cmtt}\selectfont]
     
            \highlight{gray30}{Response:}\\
        My plan is as follows:\\
        \highlight{yellow20}{[PLAN]}\\
        \tboxcomment{Provide the first action step as a hint}\\
        feast object a from object b\\
        \tboxcomment{The LLM planner starts from here}\\
        put down object 1 \tboxcomment{object 1 comes from nowhere} \\
        unstack the blue block from the orange block \\
        put down the blue block \\
        unstack the orange block from the red block \\
        put down the orange block \\
        unstack the red block from the blue block \\
        put down the red block \\
        unstack the blue block from the yellow block \\
        put down the blue block \\
        unstack the yellow block from the red block \\
        put down the yellow block \\
        unstack the orange block from the blue block \\
        put down the orange block \\
        pick up the blue block \\
        stack the blue block on top of the yellow block \\
        pick up the orange block \\
        stack the orange block on top of the blue block \\
        unstack the white block from the orange block \\
        put down the white block \\
        unstack the orange block from on top of the blue block \\
        put down the orange block \\
        unstack the blue block from on top of the yellow block \\
        put down the blue block \\
        pick up the orange block \\
        stack the orange block on top of the blue block \\
        pick up the white block \\
        stack the white block on top of the orange block  \\
        \highlight{yellow20}{[PLAN END]}
        
    \end{tcolorbox}
    \caption{We observed an interesting pattern in the responses generated by the LLM planner for obfuscated blockworld problems. The planner tends to neglect the obfuscated domain context and continue to generate the actions based on the original domain context. This behavior remains even when we provide the first action step as a hint.}
    \label{fig:app_obfuscated_response}
\end{figure}

An optimistic interpretation of this behavior is that the model is capable of 
linking the obfuscated domain back to the original domain. But eventually the model treat the obfuscated domain as an outlier and revert back to the original domain context, demonstrating that LLMs are highly sensitive to the specific vocabulary used in their training data, and struggle to generalize its planning capabilities to scenarios that deviate from its training distribution.

Two noteworthy observations emerge from the generated plan which we often call it as \emph{hallucination}:
\begin{enumerate}
    \item The model creates actions like ``put down object 1,'' where ``object 1'' doesn't exist in either the obfuscated or original problem descriptions.
    \item Interestingly, the actions generated for the original domain context often form an executable plan sequence. However, there's no one-to-one correspondence between the objects in the original and obfuscated versions.
\end{enumerate}

\section{Further Details on A Failure Case of Self-Correction Learning}
\label{app:self_correction_learning_details}
We conducted qualitative analysis to understand the effect of self-correction learning on the model's planning capabilities. In the main text, we have stated that the model is able to identify errors in a high precision and recall rate, but fails to correct them effectively. We further verify this claim by examining the generated plan sequence when the model try to solve \textsc{Blocksworld} problem in the `long' test set. The model we used is row 6 in Table~\ref{tab:ablation_study}, which is trained with the self-correction learning strategy and permutation augmentation.

\begin{figure}[H]
    \begin{minipage}{0.47\textwidth}
        
    \centering
    \begin{tcolorbox}[title=Blockworld Problem Instance Query Context, height=13.00cm, width=9cm, fontupper=\scriptsize\ttfamily\fontfamily{cmtt}\selectfont]
        \highlight{gray30}{Query:}\\
        I am playing with a set of blocks where I need to arrange the blocks into stacks.\\
        \highlight{yellow20}{[ACTION DESCRIPTION]}\\
        \emph{Here are the actions that can be performed:}\\
        (a) Pick up a block\\
        (b) Unstack a block from on top of another block\\
        (c) Put down a block\\
        (d) Stack a block on top of another block\\
        \emph{The following are the restrictions on the actions:}\\
        \dots\\
        \highlight{yellow20}{[STATEMENT]}\\
        \emph{As initial conditions I have that}\\
        the green block is on top of the red block, the yellow block is on top of the cyan block, the blue block is clear, the orange block is on top of the magenta block, the white block is on the table, the cyan block is on top of the orange block, the blue block is on top of the black block, the black block is on top of the green block, the magenta block is on top of the white block, the red block is on top of the yellow block, the hand is empty.\\

        \emph{My goal is to have that}\\
        the cyan block is on top of the red block, the blue block is on top of the cyan block, the green block is on top of the magenta block, the orange block is on top of the blue block, the red block is on top of the yellow block, the white block is on top of the green block, the yellow block is on top of the white block.\\
        
    \end{tcolorbox}
    \label{fig:app_self_correct_problem}
    \end{minipage}
    \hspace{0.5cm}
    \begin{minipage}{0.63\textwidth}
        
        \centering
        \begin{tcolorbox}[title=Response Generated By Trained LLM Planner, height=13.00cm, width=9cm, fontupper=\scriptsize\ttfamily\fontfamily{cmtt}\selectfont]
         
            \highlight{gray30}{Response:}\\
            My plan is as follows:
            \highlight{yellow20}{[PLAN]}\\
            unstack the blue block from on top of the black block\\
            put down the blue block\\
            unstack the blue block from on top of the yellow block \highlight{purple30}{WRONG}\\
            unstack the blue block from on top of the yellow block \highlight{purple30}{WRONG}\\
            unstack the blue block from on top of the blue block \highlight{purple30}{WRONG}\\
            unstack the yellow block from on top of the blue block \highlight{purple30}{WRONG}\\
            unstack the blue block from on top of the yellow block\\
            put down the blue block\\
            unstack the yellow block from on top of the blue block \highlight{purple30}{WRONG}\\
            unstack the yellow block from on top of the blue block \highlight{purple30}{WRONG}\\
            unstack the yellow block from on top of the blue block \highlight{purple30}{WRONG}\\
            unstack the blue block from on top of the blue block \highlight{purple30}{WRONG}\\
            put down the blue block\\
            unstack the yellow block from on top of the blue block \highlight{purple30}{WRONG}\\
            unstack the yellow block from on top of the blue block \highlight{purple30}{WRONG}\\
            unstack the yellow block\\
            put down the yellow block\\
            pick up the blue block\\
            stack the blue block on top of the yellow block\\
            unstack the orange block from on top of the blue block \highlight{purple30}{WRONG}\\
            unstack the blue block from on top of the yellow block\\
            put down the blue block\\
            stack the orange block on top of the blue block\\
            unstack the orange block from on top of the blue block \highlight{purple30}{WRONG}\\
            unstack the orange block from on top of the blue block \highlight{purple30}{WRONG}\\
            \dots\\
            put down the orange block\\
            unstack the blue block from on top of the yellow block\\
            put down the blue block\\
            pick up the orange block\\
            stack the orange block on top of the blue block\\
            \highlight{yellow20}{[PLAN END]}
            
        \end{tcolorbox}
        \label{fig:app_self_correct_response}
    \end{minipage}

\end{figure}

\begin{figure}[H]
    \begin{tcolorbox}[title=Blockworld Problem Instance Reference Plan, height=5.00cm, width=9cm, fontupper=\scriptsize\ttfamily\fontfamily{cmtt}\selectfont]
        (unstack blue black)  (put-down blue)  (unstack black green)  (put-down black)  (unstack green red)  (put-down green)  (unstack red yellow)  (put-down red)  (unstack yellow cyan)  (put-down yellow)  (unstack cyan orange)  (stack cyan red)  (unstack orange magenta)  (stack orange blue)  (unstack magenta white)  (put-down magenta)  (pick-up green)  (stack green magenta)  (pick-up white)  (stack white green)  (pick-up yellow)  (stack yellow white)  (unstack cyan red)  (put-down cyan)  (pick-up red)  (stack red yellow)  (pick-up cyan)  (stack cyan red)  (unstack orange blue)  (put-down orange)  (pick-up blue)  (stack blue cyan)  (pick-up orange)  (stack orange blue) 
    \end{tcolorbox}
\end{figure}

To better understand the problem, we visualize both the initial state and the goal state using the Planimation Tool \citep{chen2020planimation}. successfully recognizes the first two steps: unstacking the blue block from the black block and then placing it down. However, it fails to identify the next step. Upon examining the visualization, it becomes evident that the next action should be to unstack the black block from the green block. Instead, the model does not acknowledge the presence of the green block in the scene. This oversight is reflected in the generated plan, where interactions with the green block are notably absent, indicating that the model struggles to accurately parse the initial state. 

As a result, the LLM attempts to continue manipulating the blue block but quickly realizes this approach is incorrect, leading it to generate a \texttt{[WRONG]} token. This behavior aligns with the high precision and recall rates observed in probing tests. Nevertheless, when retrying, the model remains focused on the blue block, suggesting it is unable to effectively correct its mistake. A possible remedy for this issue is to introduce a more sophisticated mechanism that allows external expert to provide detailed feedback on why the model's action is incorrect, enabling the model to learn from its mistakes more effectively.

\begin{figure}[H]
    \begin{minipage}{0.46\textwidth}
        \centering
        \includegraphics[width=0.8\columnwidth]{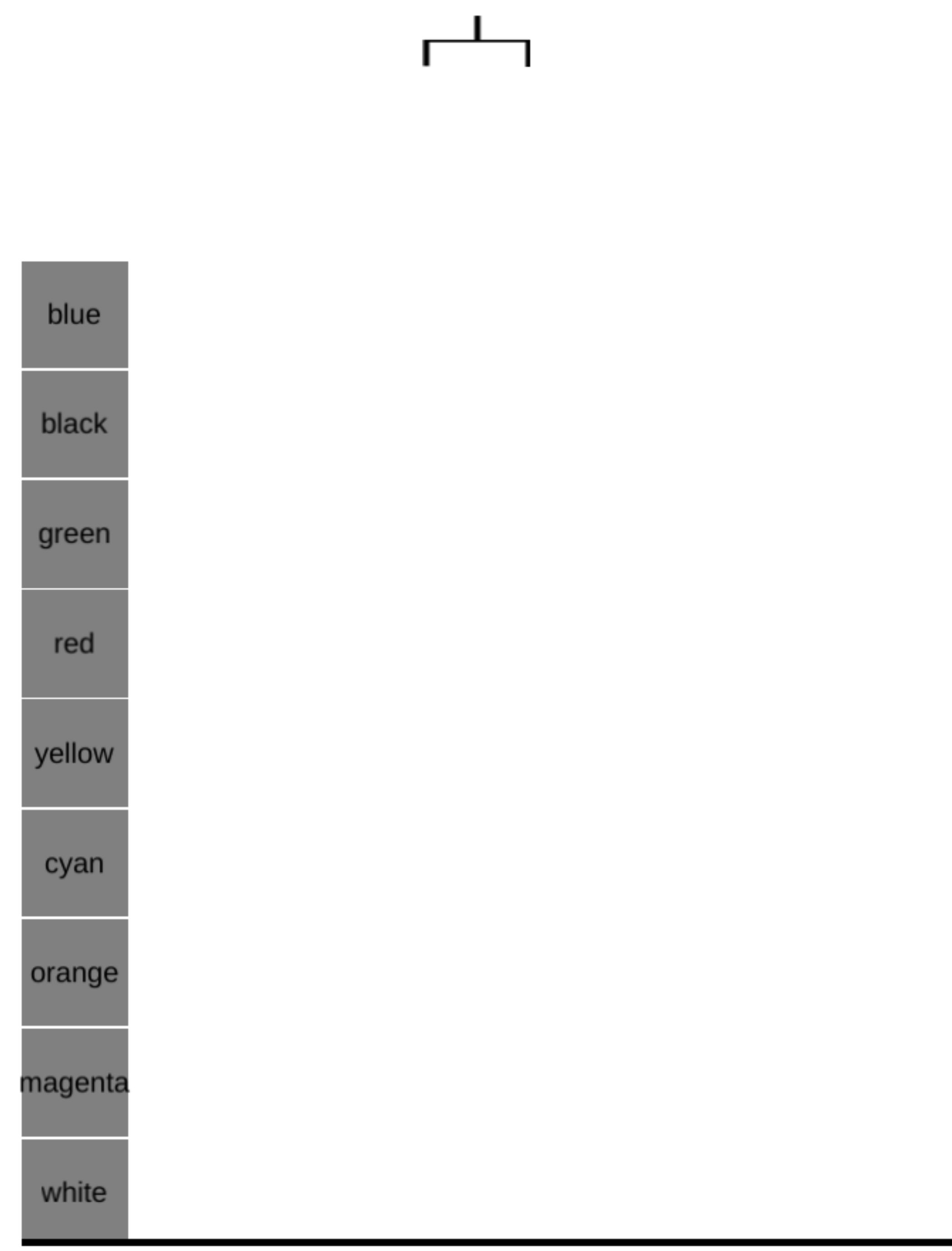}
        \caption{Initial state of the \textsc{Blocksworld} problem}
    \end{minipage}
    \hspace{1cm}
    \begin{minipage}{0.46\textwidth}
        \centering
        \includegraphics[width=0.8\columnwidth]{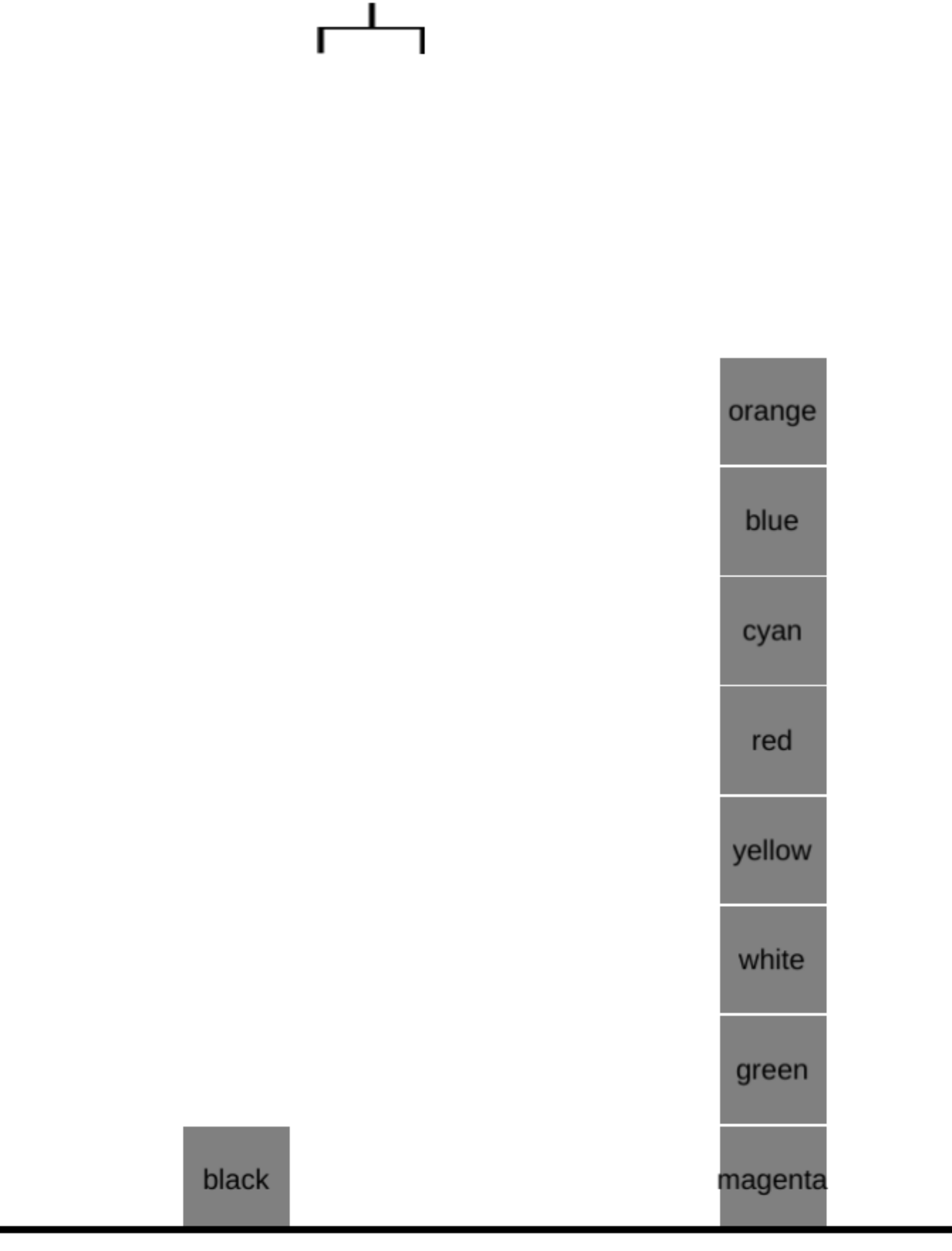}
        \caption{Goal state of the \textsc{Blocksworld} problem}
    \end{minipage}
\end{figure}

\section{Further Details on Mistake Identification Probing Tests}
\label{app:mistake_identification_probing_tests}

Probing tests utilized the feature of the autoregressive generation process inherent in LLMs, wherein each token is generated by conditioning on all preceding tokens, producing \emph{logits} scores across the vocabulary. For a greedy decoding, the model selects the token with the highest \emph{logits} score. In our case, in order to train the model to have the self-correction skills, we introduced an additional special token -- \texttt{[WRONG]} -- to the vocabulary. This token serves as an indicator for the model to recognize that the current action step is incorrect and thus re-attempt a new action. Therefore, when the model reaches the end of every generated action step, it faces a binary decision for its subsequent token: if it deems the current step correct, it generates a new line token ($\backslash n$); otherwise, it generates the \texttt{[WRONG]} token. Both options are single-token choices, ensuring a fair comparison of their conditional probabilities. 

For our probing tests, we selected the `long' test set as our evaluation dataset. We generated synthesized plan sequences using the same methodology employed in our self-correction learning strategy -- that is by randomly selecting an action that appears later in the reference plan sequence and inserting it to the current step followed by a \emph{`special removal token'}. To create instances for probing test, we processed the plans in two ways: for correct examples, we truncated the plan at the end of a correct step sentence, while for incorrect examples, we truncated at the end of an erroneous step sentence. After freezing the fine-tuned LLM, we conducted a comprehensive evaluation by measuring four distinct conditional probabilities:
\begin{enumerate}
    \item Generating the \texttt{[WRONG]} token when encountering a wrong step.
    \item Generating the new line token ($\backslash n$) when encountering a wrong step.
    \item Generating the \texttt{[WRONG]} token when encountering a correct step.
    \item Generating the new line token ($\backslash n$) when encountering a correct step.
\end{enumerate}
This 2$\times$2 matrix of probabilities allows us to assess the model's ability to discriminate between correct and incorrect steps accurately. Therefore, the precision and recall value in Table~\ref{tab:mistake_identification} is calculated based as follows:

We considered a true positive (TP) when the probability of generating the \texttt{[WRONG]} token was higher than that of generating the new line token for a wrong step, and the other way around for a correct step. The total number of wrong steps in the test set represented all actual mistakes (TN + FP). We counted an identified mistake when the probability of generating the \texttt{[WRONG]} token exceeded that of the new line token, regardless of whether the step was actually wrong or correct (TP + FP). Finally, precision was then calculated as the ratio of true positives to all identified mistakes, while recall was computed as the ratio of true positives to all actual mistakes in the test set.

\section{Additional Results: \emph{pass@k} Validity Rate}
\label{app:pass_at_k}
The \emph{pass@k} metric is an important evaluation measure for assessing the performance of LLMs in reasoning. It provides insights into the model's ability to generate correct solutions across multiple attempts. The probabilistic nature of the generative process in LLMs results in the fact that the correct plan may not always be the most confident one. Due to this, employing a majority vote over multiple sampling outputs has become a common practice to enhance the robustness of model predictions \citep{2WSLCNCZ23}. In our context, \emph{pass@k} metrics measure the validity rate by taking the best of $k$ samples generated by the model. The sampling hyperparameter are shown in Appendix~\ref{app:hyperparameters}. Results are shown in Table~\ref{tab:pass_at_k}.

\begin{figure}[H]
    \centering
    \begin{minipage}{\textwidth}
    \centering
    \captionof{table}{Pass@k validity rates for different strategies across test sets. Results show consistent improvements for `long' test sets as k increases, while `unseen' and `obfuscated' sets show no significant gains. Notably, the vanilla (1) and RL (10) strategies demonstrate the highest performance gains with multiple sampling}
    \label{tab:pass_at_k}
    \resizebox{\columnwidth}{!}{%
    \begin{tabular}{@{}c|ccccc|ccl|ccc|ccc@{}}
    \toprule
    \multirow{2}{*}{Row} & \multicolumn{5}{c|}{\textbf{Strategies}}                                                                                                                                                                         & \multicolumn{3}{c|}{\textbf{Long}}                     & \multicolumn{3}{c|}{\textbf{Unseen}}                      & \multicolumn{3}{c}{\textbf{Obfuscated}}                  \\ \cmidrule(l){2-15} 
                           & Perm.                           & \begin{tabular}[c]{@{}c@{}}Goal\\ CoT\end{tabular} & \begin{tabular}[c]{@{}c@{}}State\\ CoT\end{tabular} & \begin{tabular}[c]{@{}c@{}}Self\\ Correct\end{tabular} & RL         & pass@1 & pass@3          & pass@5                      & pass@1       & pass@3       & \multicolumn{1}{l|}{pass@5} & pass@1       & pass@3       & \multicolumn{1}{l}{pass@5} \\ \midrule
    \textbf{1$\star$}      & \multicolumn{1}{c|}{}           & \multicolumn{1}{c|}{}                              & \multicolumn{1}{c|}{}                               & \multicolumn{1}{c|}{}                                  &            & 34.8\% & \goodval 43.9\% & \goodval 49.0\%             & \textbf{0\%} & 12.2\%       & 12.2\%                      & \textbf{0\%} & \textbf{0\%} & \textbf{0\%}               \\ \midrule
    2                      & \multicolumn{1}{c|}{\checkmark} & \multicolumn{1}{c|}{}                              & \multicolumn{1}{c|}{}                               & \multicolumn{1}{c|}{}                                  &            & 35.0\% & 40.4\%          & 42.9\%                      & \textbf{0\%} & \textbf{0\%} & \textbf{0\%}                & \textbf{0\%} & \textbf{0\%} & \textbf{0\%}               \\ \midrule
    3                      & \multicolumn{1}{c|}{\checkmark} & \multicolumn{1}{c|}{\checkmark}                    & \multicolumn{1}{c|}{}                               & \multicolumn{1}{c|}{}                                  &            & 12.1\% & 15.6\%          & 17.8\%                      & 5.5\%        & 6.0\%        & \multicolumn{1}{l|}{7.0\%}  & \textbf{0\%} & \textbf{0\%} & \textbf{0\%}               \\ \midrule
    4                      & \multicolumn{1}{c|}{\checkmark} & \multicolumn{1}{c|}{}                              & \multicolumn{1}{c|}{\checkmark}                     & \multicolumn{1}{c|}{}                                  &            & 29.5\% & 35.3\%          & 37.2\%                      & \textbf{0\%} & \textbf{0\%} & \textbf{0\%}                & \textbf{0\%} & \textbf{0\%} & \textbf{0\%}               \\ \midrule
    5                      & \multicolumn{1}{c|}{\checkmark} & \multicolumn{1}{c|}{\checkmark}                    & \multicolumn{1}{c|}{\checkmark}                     & \multicolumn{1}{c|}{}                                  &            & 23.8\% & 28.4\%          & \multicolumn{1}{c|}{34.1\%} & \textbf{0\%} & \textbf{0\%} & \textbf{0\%}                & \textbf{0\%} & \textbf{0\%} & \textbf{0\%}               \\ \midrule
    6                      & \multicolumn{1}{c|}{\checkmark} & \multicolumn{1}{c|}{}                              & \multicolumn{1}{c|}{}                               & \multicolumn{1}{c|}{\checkmark}                        &            & 32.6\% & 40.4\%          & 42.3\%                      & \textbf{0\%} & \textbf{0\%} & \textbf{0\%}                & \textbf{0\%} & \textbf{0\%} & \textbf{0\%}               \\ \midrule
    7                      & \multicolumn{1}{c|}{\checkmark} & \multicolumn{1}{c|}{\checkmark}                    & \multicolumn{1}{c|}{}                               & \multicolumn{1}{c|}{\checkmark}                        &            & 14.9\% & 21.0\%          & 24.3\%                      & \textbf{0\%} & \textbf{0\%} & \textbf{0\%}                & \textbf{0\%} & \textbf{0\%} & \textbf{0\%}               \\ \midrule
    8                      & \multicolumn{1}{c|}{\checkmark} & \multicolumn{1}{c|}{}                              & \multicolumn{1}{c|}{\checkmark}                     & \multicolumn{1}{c|}{\checkmark}                        &            & 27.5\% & 31.4\%          & 35.1\%                      & \textbf{0\%} & \textbf{0\%} & \textbf{0\%}                & \textbf{0\%} & \textbf{0\%} & \textbf{0\%}               \\ \midrule
    9                      & \multicolumn{1}{c|}{\checkmark} & \multicolumn{1}{c|}{\checkmark}                    & \multicolumn{1}{c|}{\checkmark}                     & \multicolumn{1}{c|}{\checkmark}                        &            & 25.9\% & 31.2\%          & \multicolumn{1}{c|}{36.5\%} & \textbf{0\%} & \textbf{0\%} & \textbf{0\%}                & \textbf{0\%} & \textbf{0\%} & \textbf{0\%}               \\ \midrule
    \textbf{10$\star$}     & \multicolumn{1}{c|}{}           & \multicolumn{1}{c|}{}                              & \multicolumn{1}{c|}{}                               & \multicolumn{1}{c|}{}                                  & \checkmark & 41.5\% & \goodval 49.2\% & \goodval 52.0\%             & 12.5\%       & 12.5\%       & 12.5\%                      & \textbf{0\%} & \textbf{0\%} & \textbf{0\%}               \\ \midrule
    11                     & \multicolumn{1}{c|}{}           & \multicolumn{1}{c|}{}                              & \multicolumn{1}{c|}{}                               & \multicolumn{1}{c|}{\checkmark}                        & \checkmark & 36.3\% & 42.3\%          & \multicolumn{1}{c|}{45.0\%} & \textbf{0\%} & \textbf{0\%} & \textbf{0\%}                & \textbf{0\%} & \textbf{0\%} & \textbf{0\%}               \\ \bottomrule
    \end{tabular}%
    }
\end{minipage}
\end{figure}

The results indicate that multiple sampling does not enhance the model's performance on the `unseen' and `obfuscated' domains, highlighting the model's limited generalization to unfamiliar contexts. In contrast, we observed a consistent improvement in the \emph{pass@k} validity rate across all strategies for the `long' test set. Among all the ablation strategies, the vanilla (row 1) and RL (row 10) models demonstrate the most significant improvements when utilizing multiple sampling. So, we actually observe this trend -- strategies that increases the response length, such as incorporating Chain of Thought prompts (row 3 to 9), actually cannot benefit from multiple sampling. We attribute this phenomenon to the nature of autoregressive prediction in LLMs. That is, as the response length increases, the context provided by all preceding tokens becomes denser and more concrete. By conditioning on this context, the model will have more focused and less diverse set of plausible continuations for the next token. Consequently, the benefits of generating multiple samples are diminished for most strategies.

\section{Additional Results: Goal Satisfiability Rate}
\label{app:goal_satisfiability_result}
It is important to note that goal satisfiability is also not a standard term in the planning literature. Nevertheless, an informal definition of goal satisfiability, derived from the definition of executability, is as follows:

\begin{definition}[Goal-Satisfiability of a Plan]
    A plan is goal-satisfiable if it defines an action sequence $(a_0, a_1, \ldots, a_{n-1})$ with states $(s_0, s_1, \ldots, s_n)$. $s_0$ is the initial state and for each $i=0, 1, \ldots, n-1$, $s_{i+1}$ is the result of executing $a_i$ in $s_i$, removing delete effects, adding add effects and applying numeric effects. The state $s_n$ is called the final goal state and the goal $G$ holds in $s_n$. A goal-satisfiable plan may not be executable.
\end{definition}

We also measured the \emph{goal satisfiability rate} in the interest of completeness. However, there's a fundamental misalignment between this metric and the nature of autoregressive language models in end-to-end plan generation. Here's why:

\begin{itemize}
    \item The sequential nature of autoregressive language models allows them to generate plans one token at a time, moving from left to right, similar to the forward progression of a plan sequence.
    \item Each new prediction by nature aims to maintain consistency with preceding ones.
    \item This inherently prioritizes local state transition coherence over goal satisfaction, just like how forward progression in a plan sequence will not jump to the goal state without ensuring the coherence of the preceding actions.
    \item Existing strategies, particularly \emph{State CoT}, often emphasize the consistency of the local step transitions, therefore, pursuing goal satisfiability before ensuring executability conflicts with the idea of producing a plan sequence in a left-to-right manner.
\end{itemize}

Therefore, we say that the goal satisfiability metric fail to account for the characteristics of end-to-end plan generation in autoregressive language models. Nevertheless, this can be also seen as a limitation of the end-to-end left-to-right plan generation paradigm, as the trained model lack mechanisms for looking ahead to future states and conducting backward searches. It will eventually undermines the model's ability to generate a valid plan. 

The results of the goal satisfiability rate are shown in Table~\ref{tab:goal_satisfiability_result}.

\begin{figure}[H]
    \centering
    \begin{minipage}{0.7\textwidth}
        
    \centering
    \captionof{table}{The results show a slight decrease in the goal satisfiability rate across the different strategies. This indicates that the strategies do focus on the sequential consistency due to the nature of autoregressive models and prioritize less on the goal satisfiability.}
    \label{tab:goal_satisfiability_result}

    \resizebox{\textwidth}{!}{%
    \begin{tabular}{@{}c|ccccc|c|c|c|c@{}}
    \toprule
    \multirow{2}{*}{Row} & \multicolumn{5}{c|}{\textbf{Strategies}}                                                                                                                                                                         & \textbf{In-Distrib.} & \textbf{Long}   & \textbf{Unseen} & \textbf{Obfuscated} \\ \cmidrule(l){2-10} 
                           & Perm.                           & \begin{tabular}[c]{@{}c@{}}Goal\\ CoT\end{tabular} & \begin{tabular}[c]{@{}c@{}}State\\ CoT\end{tabular} & \begin{tabular}[c]{@{}c@{}}Self\\ Correct\end{tabular} & RL         & goal sat.            & goal sat.       & goal sat.       & goal sat.           \\ \midrule
    \textbf{1$\star$}      & \multicolumn{1}{c|}{}           & \multicolumn{1}{c|}{}                              & \multicolumn{1}{c|}{}                               & \multicolumn{1}{c|}{}                                  &            & \textbf{100\%}       & 64.5\%          & \goodval 63.5\% & \textbf{0\%}        \\ \midrule
    2                      & \multicolumn{1}{c|}{\checkmark} & \multicolumn{1}{c|}{}                              & \multicolumn{1}{c|}{}                               & \multicolumn{1}{c|}{}                                  &            & \textbf{100\%}       & 62.0\%          & 25.5\%          & \textbf{0\%}        \\ \midrule
    3                      & \multicolumn{1}{c|}{\checkmark} & \multicolumn{1}{c|}{\checkmark}                    & \multicolumn{1}{c|}{}                               & \multicolumn{1}{c|}{}                                  &            & \textbf{100\%}       & \badval 35.0\%  & 27.0\%          & \textbf{0\%}        \\ \midrule
    4                      & \multicolumn{1}{c|}{\checkmark} & \multicolumn{1}{c|}{}                              & \multicolumn{1}{c|}{\checkmark}                     & \multicolumn{1}{c|}{}                                  &            & \textbf{100\%}       & 57.0\%          & \textbf{0\%}    & \textbf{0\%}        \\ \midrule
    5                      & \multicolumn{1}{c|}{\checkmark} & \multicolumn{1}{c|}{\checkmark}                    & \multicolumn{1}{c|}{\checkmark}                     & \multicolumn{1}{c|}{}                                  &            & 97.5\%               & 55.5\%          & 8.0\%           & \textbf{0\%}        \\ \midrule
    6                      & \multicolumn{1}{c|}{\checkmark} & \multicolumn{1}{c|}{}                              & \multicolumn{1}{c|}{}                               & \multicolumn{1}{c|}{\checkmark}                        &            & \textbf{100\%}       & 57.5\%          & 28.5\%          & \textbf{0\%}        \\ \midrule
    7                      & \multicolumn{1}{c|}{\checkmark} & \multicolumn{1}{c|}{\checkmark}                    & \multicolumn{1}{c|}{}                               & \multicolumn{1}{c|}{\checkmark}                        &            & 96.0\%               & 46.5\%          & 30.5\%          & \textbf{0\%}        \\ \midrule
    8                      & \multicolumn{1}{c|}{\checkmark} & \multicolumn{1}{c|}{}                              & \multicolumn{1}{c|}{\checkmark}                     & \multicolumn{1}{c|}{\checkmark}                        &            & \textbf{100\%}       & 56.5\%          & \textbf{0\%}    & \textbf{0\%}        \\ \midrule
    9                      & \multicolumn{1}{c|}{\checkmark} & \multicolumn{1}{c|}{\checkmark}                    & \multicolumn{1}{c|}{\checkmark}                     & \multicolumn{1}{c|}{\checkmark}                        &            & 98.5\%               & 53.5\%          & \textbf{0\%}    & \textbf{0\%}        \\ \midrule
    \textbf{10$\star$}     & \multicolumn{1}{c|}{}           & \multicolumn{1}{c|}{}                              & \multicolumn{1}{c|}{}                               & \multicolumn{1}{c|}{}                                  & \checkmark & 98.5\%               & \goodval 73.5\% & \goodval 53.5\% & \textbf{0\%}        \\ \midrule
    11                     & \multicolumn{1}{c|}{}           & \multicolumn{1}{c|}{}                              & \multicolumn{1}{c|}{}                               & \multicolumn{1}{c|}{\checkmark}                        & \checkmark & \textbf{100\%}       & 60.0\%          & 29.0\%          & \textbf{0\%}        \\ \bottomrule
    \end{tabular}%
    }
\end{minipage}

\end{figure}

We observed a decrease in the goal satisfiability rate across various strategies. This result is reasonable, as improvements in both the executability and goal satisfiability rates would naturally lead to an increase in the final validity rate, which is not the case in our experiments. As shown in Table~\ref{tab:goal_satisfiability_result}, the vanilla model (row 1) achieved a high goal satisfiability rate across the test sets, though this did not correspond to a high validity rate, as indicated by our experiments discussed in the main text. The evaluated strategies maintained a considerable invariance in the goal satisfiability rate for the `long' test set but experience a decrease in the `unseen' test set. Specifically, the \emph{State CoT} strategy shows the most significant drop, primarily due to its focus on maintaining the state transition consistency of the generated plans.

\end{document}